\documentclass[journal=iecred,manuscript=article]{achemso}
\setkeys{acs}{doi = true,maxauthors=3}

\usepackage{chemformula}
\usepackage[version=3]{mhchem} 
\usepackage{longtable} 
\usepackage{pdflscape} 
\usepackage{multirow}
\usepackage{placeins}
\usepackage{hyperref}
\usepackage{subcaption}
\usepackage{bm}
\usepackage{booktabs}
\usepackage{lineno}

\newcommand{\Newnameref}[1]{\textit{\nameref{#1}}}

\author{Thomas Specht}
\affiliation{Laboratory of Engineering Thermodynamics (LTD), RPTU Kaiserslautern, Germany}
\author{Mayank Nagda}
\affiliation{Department of Computer Science, RPTU Kaiserslautern, Germany}
\author{Sophie Fellenz}
\affiliation{Department of Computer Science, RPTU Kaiserslautern, Germany}
\author{Stephan Mandt}
\affiliation{Department of Computer Science, University of California, Irvine, CA, USA}
\author{Hans Hasse}
\affiliation{Laboratory of Engineering Thermodynamics (LTD), RPTU Kaiserslautern, Germany}
\author{Fabian Jirasek}
\email{*fabian.jirasek@rptu.de}
\affiliation{Laboratory of Engineering Thermodynamics (LTD), RPTU Kaiserslautern, Germany}

\title{HANNA: Hard-constraint Neural Network for Consistent Activity Coefficient Prediction}

\begin{document}



\begin{abstract}
We present the first hard-constraint neural network for predicting activity coefficients (HANNA), a thermodynamic mixture property that is the basis for many applications in science and engineering. Unlike traditional neural networks, which ignore physical laws and result in inconsistent predictions, our model is designed to strictly adhere to all thermodynamic consistency criteria. By leveraging deep-set neural networks, HANNA maintains symmetry under the permutation of the components. Furthermore, by hard-coding physical constraints in the network architecture, we ensure consistency with the Gibbs-Duhem equation and in modeling the pure components. The model was trained and evaluated on 317,421 data points for activity coefficients in binary mixtures from the Dortmund Data Bank, achieving significantly higher prediction accuracies than the current state-of-the-art model UNIFAC. Moreover, HANNA only requires the SMILES of the components as input, making it applicable to any binary mixture of interest. HANNA is fully open-source and available for free use.
\end{abstract}
\section{Introduction}
Neural networks (NNs) have recently revolutionized many fields, including image analysis~\cite{Ravindran2022}, speech recognition~\cite{LeCun2015}, predicting protein folding~\cite{Jumper2021, Abramson2024}, and language modeling~\cite{MBran2024, VanVeen2024}. These models are universal and highly flexible function approximators~\cite{Hornik1989}, which perform best if they have large amounts of training data. NNs are also gaining more and more attention in chemical engineering~\cite{Venkatasubramanian2018,Fang2022, Li2022, Zang2023,Schweidtmann2024} but face two significant challenges preventing them from exploiting their full potential in this field: sparse training data and inconsistent predictions. Like in other fields of science and engineering, data sparsity is ubiquitous in chemical engineering due to the high effort and costs related to experimental data collection, making predictions with purely data-driven NNs difficult. Furthermore, since NNs are a priori agnostic about physical laws and boundaries, there is no guarantee that their predictions obey these rules, frequently leading to physically inconsistent results and predictions~\cite{Rittig2023}. This, in turn, is detrimental to the trust in NN-based models and a severe obstacle to their adoption and use in practice.

The most promising solution to these challenges is to incorporate explicit physical knowledge into NNs to support their training beyond using only the limited available data. Most prominently, Physics-Informed Neural Networks (PINNs)~\cite{Raissi2019} have been successfully applied in different fields~\cite{Li2022, Lin2022, Zhu2021, Molnar2022, Psaros2021, Martin2022, Zhao2022, Rittig2023}, primarily to solve partial differential equations (PDE) efficiently. PINNs incorporate the governing physical equation or boundary conditions into the loss function of an NN by adding a term that penalizes solutions deviating from the constraint (e.g., the compliance of a PDE)~\cite{Karniadakis2021}. PINNs are inherently soft-constraint methods that do not enforce \textit{exact} compliance with the given constraints, which is a well-known limitation of penalty methods in general~\cite{Xu2020, Chen2021} and has potential drawbacks. Specifically, while approximately complying with physical laws and boundaries might be sufficient in some cases, this is unacceptable in many applications; for instance, thermodynamic models that yield physically inconsistent predictions will not be accepted and used in chemical engineering practice.

Hard-constraint models, which strictly enforce physical relations and constraints in NNs, are generally considered challenging to develop~\cite{Karniadakis2021, Chen2021, Chen2020,Neila2017, Lu2021}. Thermodynamics is the ideal field for designing such hard-constraint models with its extensive treasure of explicit physical knowledge on the one hand and the high demand for strict compliance of predictive thermodynamic models with physical laws and constraints on the other.  In this work, we introduce the first hard-constraint NN-based model for thermodynamic property prediction, which opens up an entirely new way of thermodynamic model development but also holds the promise to advance model development in other fields of chemical engineering and beyond.

Predicting the thermodynamic properties of pure components and mixtures is fundamental in many fields of science and engineering. In chemical engineering, knowledge of thermodynamic properties is the basis for process design and optimization. However, experimental data on thermodynamic properties are scarce. The problem is particularly challenging for mixtures, where missing data are prevalent due to the combinatorial complexity involved.

One of the most critical thermodynamic properties is the activity coefficient of a component in a mixture. Activity coefficients are central for calculating reaction and phase equilibria and, therefore, the basis for modeling and simulating chemical processes. Since activity coefficients cannot be measured directly, they are usually determined indirectly by evaluating phase equilibrium experiments. Since these experiments are time-consuming and expensive, experimental data on activity coefficients are often lacking, and many physical prediction methods have been developed and are widely applied in industry~\cite{Gmehling2019-vx}.

Physical methods for predicting activity coefficients model the molar Gibbs excess energy $g^\text{E}$ as a function of temperature $T$ and mixture composition in mole fractions $\bm{x}$, from which the logarithmic activity coefficients $\text{ln} \gamma_i$ are obtained by partial differentiation~\cite{Gmehling2019-vx}. The two most widely used $g^\text{E}$ models are NRTL~\cite{Renon1968} and UNIQUAC~\cite{Abrams1975}. These models generalize over state points, i.e., temperature and mole fractions, but cannot extrapolate to unstudied mixtures. In contrast, $g^\text{E}$ models based on quantum-chemical descriptors, such as COSMO-RS~\cite{Klamt1995} and COSMO-SAC~\cite{Lin2001,Hsieh2010, Hsieh2014}, or group contributions models, such as the different versions of UNIFAC~\cite{Fredenslund1975,Weidlich1987} (with modified UNIFAC (Dortmund) being the most advanced \cite{Weidlich1987}) also allow to generalize over components and mixtures. However, even though they have been continuously developed and refined for decades, the state-of-the-art models show significant weaknesses for certain classes of components. The consequential inaccuracies in predicting activity coefficients result in wrongly predicted phase equilibria, leading to poor process modeling and simulation~\cite{Fingerhut2017, Jirasek2020b}. On the upside, the theoretical foundation of the established physical models allows for good extrapolation performance, and, even more importantly, they exhibit strict compliance with thermodynamic laws, boundaries, and consistency criteria. 

Recently, machine learning (ML) methods have gained attention for predicting activity coefficients~\cite{Jirasek2020a, Damay2023,SanchezMedina2022} and other thermodynamic properties~\cite{Santana2024,Habicht2023,Deng2023, Shilpa2023,Aouichaoui2023, Hayer2022, Gromann2022}. Even though these models are purely data-driven, they surpassed the physical thermodynamic models in prediction accuracy. However, they were all limited to specific state points and could, e.g., not describe the composition dependence of activity coefficients. 

To improve the ML models further, various hybridization approaches~\cite{Jirasek2023Review} were developed that combine the flexibility of ML methods with physical knowledge. This was, e.g., done by augmenting the training data with synthetic data obtained from physical prediction methods~\cite{Jirasek2020b, Winter2022}. Other recently developed hybridization approaches~\cite{Jirasek2022, Jirasek2023_MCM, Winter2023} have broadened the application range of physical thermodynamic models. In these approaches, an ML method is embedded in a physical thermodynamic model to predict the parameters of the physical model. By retaining the framework of the physical models, these hybrid models are intrinsically thermodynamically consistent. On the downside, these models are subject to the same assumptions and simplifications taken during the development of the original model, limiting their flexibility. Consequently, they have a restricted value range of predictable activity coefficients~\cite{Werner2023}, limiting the description of certain phase behaviours~\cite{Rarey2005, Marcilla2011,Marcilla2018,Marcilla2019}. 

Rittig et al. \cite{Rittig2023} recently developed a PINN taking thermodynamic constraints into account to predict activity coefficients in binary mixtures; however, their study was limited to synthetic data, and physical information was only included with a soft constraint, i.e., without strictly enforcing thermodynamic consistency. Hybrid models for activity coefficient prediction that \textit{fully} exploit the flexibility of NNs while \textit{guaranteeing} thermodynamic consistency have not been available until now. This work has addressed this gap.

Specifically, we have developed the first hard-constraint NN for the Gibbs excess energy $g^\text{E}$ of a mixture, which allows us to predict activity coefficients $\text{ln} \gamma_i$ in any binary mixture of arbitrary components at any state point. We name our method \underline{HA}rd-constraint \underline{N}eural \underline{N}etwork for \underline{A}ctivity coefficient prediction (HANNA) in the following. We restrict ourselves here to binary mixtures, which is inconsequential from a practical standpoint. All physical models of mixtures are based on pair interactions, which can, and practically always are, trained on data for binary mixtures. The physical foundation then enables extrapolations to multicomponent mixtures (with an arbitrary number of components) in a purely predictive manner. Since HANNA explicitly considers all relevant physical knowledge, it can also be easily extended to multicomponent mixtures.

The Gibbs excess energy $g^\text{E}$ of a mixture, and consequently the activity coefficients $\text{ln} \gamma_i$, are typically expressed as functions of temperature $T$, pressure $p$, and the composition in mole fractions $\bm{x}$ of the components. In the following, we will express $g^\text{E}$ and the activity coefficients $\text{ln} \gamma_i$ in binary mixtures as functions of $T$, $p$, and $x_1$. For liquid mixtures, the influence of the pressure is small and is often neglected, which is also the case for our model. However, for the sake of clarity, all thermodynamic derivations are written here without this assumption.

The predictions of HANNA strictly comply with all relevant thermodynamic consistency criteria, which are listed for binary mixtures as follows.

\begin{enumerate}
    \item[1)] The activity coefficients of pure components are unity: 
        \begin{equation}
        \lim_{x_i \to 1} \text{ln} \gamma_i (T,p,x_i) = 0
        \label{unity}\end{equation}
    \item[2)] The activity coefficients of the components in a mixture are coupled by the Gibbs-Duhem equation, which reads for the binary mixture:
        \begin{equation}
         x_1\left (\frac{\partial \text{ln}\gamma_{1} }{\partial x_1}  \right )_{T,p}+ (1-x_1)\left (\frac{\partial \text{ln}\gamma_{2} }{\partial x_1}  \right )_{T,p}=0
        \label{Gibbs-Duhem}
        \end{equation}
    \item[3)] The activity coefficients in a pseudo-binary mixture $\mathrm{A}+\mathrm{B}$ where $\mathrm{A}=\mathrm{B}$ are always unity: 
        \begin{equation}
        \text{ln} \gamma_i(T,p,x_i)=0
        \label{unity_pseudo}
        \end{equation}
    \item[4)] Upon changing the order of the components in the input of a model for predicting the activity coefficients $\text{ln}\gamma_{1}$ and $\text{ln}\gamma_{2}$ in a binary mixture, the values of the predicted activity coefficients must not change, only their order. Mathematically, this is called permutation-equivariance and can be expressed as: 
    \begin{equation}
        \bm{\gamma}(P(\bm{x})) = P(\bm{\gamma}(\bm{x}))
        \label{Equivariance}
    \end{equation}
    where $\bm{\gamma}$ is the vector containing the (logarithmic) activity coefficients of the mixture components, $\bm{x}$ is the vector containing the information on the components in the input, including their descriptors and mole fractions, and $P$ is a permutation operator.
\end{enumerate}

Equations~\eqref{unity} and \eqref{Equivariance} also hold for multicomponent mixtures, whereas Equations~\eqref{Gibbs-Duhem} and \eqref{unity_pseudo} can easily be generalized to multicomponent mixtures.

In Figure~\ref{Overview}, we visualize how HANNA strictly enforces these constraints for predicting activity coefficients, leading to the novel class of hybrid NNs developed in this work. The central idea is to learn the molar excess Gibbs energy $g^\text{E}$ of the mixture rather than the individual activity coefficients ($\gamma_1$ and $\gamma_2$) directly. The values of $\gamma_1$ and $\gamma_2$ can then be obtained from $g^\text{E}$ by the laws of thermodynamics, ensuring strict thermodynamic consistency. HANNA consists of four parts:
\begin{figure}[t]
\centering
 \includegraphics[scale=0.62]{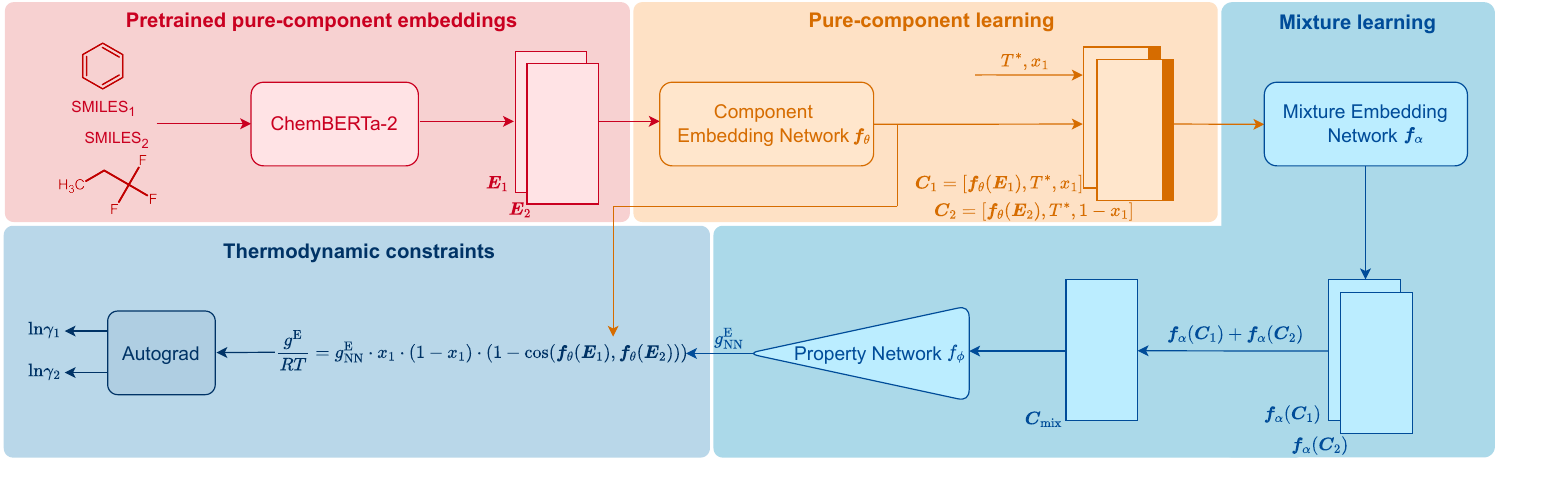}
\caption{Scheme of HANNA, the first hard-constraint NN for predicting activity coefficients in binary mixtures. Technical details on the architecture are given in Section~\Newnameref{Data_Splitting_Training}.}
\label{Overview}
\end{figure}
\FloatBarrier

\begin{enumerate}
    \item[\textbf{1)}] \textbf{Pure-component embeddings from pretrained ChemBERTa-2}\\
    We use SMILES~\cite{Weininger1988} strings to represent the components and preprocess them with ChemBERTa-2~\cite{ahmad2022}, a language model pretrained on an extensive database of molecules for learning ``pure component embeddings'' of the molecules from the respective SMILES.
    \item[\textbf{2)}] \textbf{Refining pure-component embeddings for thermodynamic property prediction}\\
      Since the embeddings of ChemBERTa-2 were not explicitly trained on thermodynamic properties, we ``fine-tune'' them to predict thermodynamic properties in a two-step process. We first feed them into a ``component embedding network'' $\bm{f}_\theta$ to get a lower dimensional representation of each component $i$. Then, the information on the standardized temperature $T^*$ (see ~Section~\Newnameref{Data_Splitting_Training} for the definition) and the composition (here: mole fraction $x_1$ of component 1) are concatenated to each of the component embeddings. The result of this step is a refined embedding for each component $i$, represented as vector $\bm{C}_i$,  tailored for thermodynamic mixture property prediction. 
    \item[\textbf{3)}] \textbf{Learning mixture embeddings and preliminarly prediction}\\
      The component embeddings $\bm{C}_i$ are then individually processed by the ``mixture embedding network'' $\bm{f}_\alpha$, whose outputs are then aggregated using the sum operation to yield $\bm{C}_{\text{mix}}$. This step guarantees permutation invariance, i.e., independence of the order of the components, an idea inspired by deep-set models~\cite{Zaheer2017, Hanaoka2020}, and ensures that Equation~\eqref{Equivariance} is fulfilled. Subsequently, the sum is fed into another ``property prediction'' network $f_\phi$ whose output $g^\text{E}_{\text{NN}}$ is a scalar that can be understood as a preliminary prediction of the molar Gibbs excess energy $g^\text{E}$ of the mixture.
    \item[\textbf{4)}] \textbf{Enforcing all physical consistency criteria}\\
    In this step, $g^\text{E}_{\text{NN}}$ is further processed to guarantee the compliance of HANNA's predictions with the remaining consistency criteria, cf. Equations~\eqref{unity}-\eqref{unity_pseudo}. Step 4 basically corrects the preliminary $g^\text{E}_{\text{NN}}$ to hard-constrain the final predicted molar Gibbs excess energy $g^\text{E}$ on physically consistent solutions. Specifically, $g^\text{E}$ of the mixture of interest is calculated by:
    \begin{equation}
    \frac{g^\text{E}}{RT}=g^\text{E}_{\text{NN}}\cdot x_1 \cdot (1-x_1) \cdot (1-\cos(\bm{f}_\theta(\bm{E}_1), \bm{f}_\theta(\bm{E}_2)))
    \label{ge_HNN}
    \end{equation}
    where
    \begin{equation}
    1-\cos(\bm{f}_\theta(\bm{E}_1), \bm{f}_\theta(\bm{E}_2))=1-\frac{\bm{f}_\theta(\bm{E}_1)\cdot \bm{f}_\theta(\bm{E}_2)}{\left \| \bm{f}_\theta(\bm{E}_1) \right \|_2\left \| \bm{f}_\theta(\bm{E}_2)  \right \|_2} 
    \label{ge_cosine}
    \end{equation}
    denotes the cosine distance between the two component embeddings $\bm{f}_\theta(\bm{E}_1)$ and $ \bm{f}_\theta(\bm{E}_2)$, $R$ is the ideal gas constant, and $T$ is the absolute temperature. The term $x_1 \cdot (1-x_1)$ in Equation~\eqref{ge_HNN} ensures that $g^\text{E}$ becomes zero in the case of pure components ($x_1=1$ or $x_1=0$), thereby enforcing strict consistency with regard to Equation~\eqref{unity}. The cosine distance, cf. Equation~\eqref{ge_cosine}, ensures that if the two component embeddings are identical, i.e., the studied ``mixture'' is, in fact, a pure component (cosine distance equals zero), $g^\text{E}$ always becomes zero to guarantee consistency regarding Equation~\eqref{unity_pseudo}.
    
    Finally, the logarithmic activity coefficients $\text{ln}\gamma_i$ are derived in a thermodynamically consistent way from $g^\text{E}$ by partial differentiation, which reads for a binary mixture~\cite{Gmehling2019-vx,Deiters2012-me}:
    \begin{equation}
    \begin{aligned}
        \text{ln}\gamma_{1} &= \frac{g^\text{E}}{RT}+(1-x_1)\frac{\left(\frac{\partial g^\text{E}}{\partial x_1}\right)_{T,p}}{RT}  \\
        \text{ln}\gamma_{2} &= \frac{g^\text{E}}{RT}-x_1\frac{\left(\frac{\partial g^\text{E}}{\partial x_1}\right)_{T,p}}{RT}
    \end{aligned}
    \label{gammafromgibbs}
\end{equation}

    For this purpose, the auto-differentiation function ``autograd'' from pytorch~\cite{NEURIPS2019_9015} is used to calculate  $\text{ln}\gamma_i$ following Equation~\eqref{gammafromgibbs}. This last step intrinsically ensures the Gibbs-Duhem consistency of the predicted activity coefficients, cf.~Equation~\eqref{Gibbs-Duhem}. Furthermore, since $g^\text{E}$ is enforced to be permutation-invariant in step 3, the differentiation in Equation~\eqref{gammafromgibbs} always yields permutation-equivariant predictions for $\text{ln}\gamma_i$.

\end{enumerate}

HANNA was trained end-to-end and evaluated on 317,421 data points for $\text{ln}\gamma_i$ in 35,012 binary systems from the Dortmund Data Bank (DDB)~\cite{DDB2023}, cf. Section~\Newnameref{Data} for details. The data set was randomly split system-wise in 80\% training, 10\% validation, and 10\% test set. Technical details on HANNA and the optimization procedure are given in Section~\Newnameref{Data_Splitting_Training}. We also trained and validated a version of HANNA on 100\% of the data with the final set of hyperparameters. This version is not discussed or used to evaluate the predictive performance of HANNA in this work but will be provided together with this paper as an open-source version. This final version of HANNA should be used if activity coefficients in any binary mixture need to be predicted. The only inputs needed are the SMILES of the components, their mole fractions, and the temperature.
\FloatBarrier

\section{Results}

In the following, we discuss the performance of HANNA for predicting activity coefficients from the test set, which were not used for training or hyperparameter optimization. For comparison, we also include the results of modified UNIFAC (Dortmund)~\cite{Weidlich1987}, referred to simply as UNIFAC in the following. The UNIFAC training set has not been disclosed. However, since the groups developing UNIFAC and maintaining the DDB are essentially the same, one can assume that a large share of the data considered here was also used for training UNIFAC. Hence, contrary to the results of HANNA, the results obtained with UNIFAC cannot be considered true predictions. This generates a strong bias of the comparison in favor of UNIFAC.

We compare the performance of the models using a system-wise error score. Specifically, we calculate system-specific mean absolute errors (MAE) by averaging the absolute deviations of the predicted logarithmic activity coefficients from the experimental data for each system from the test set. This procedure ensures equal weighting of all systems irrespective of the number of data points and prevents overweighting well-studied systems like water + ethanol. All 3,502 systems in the test set can be predicted with HANNA, but due to missing parameters, only 1,658 can be modeled with UNIFAC. Therefore, both models are compared on the smaller shared horizon, called the ``UNIFAC horizon'' in the following.

\begin{figure}[ht]
\centering
 \includegraphics[scale=0.53]{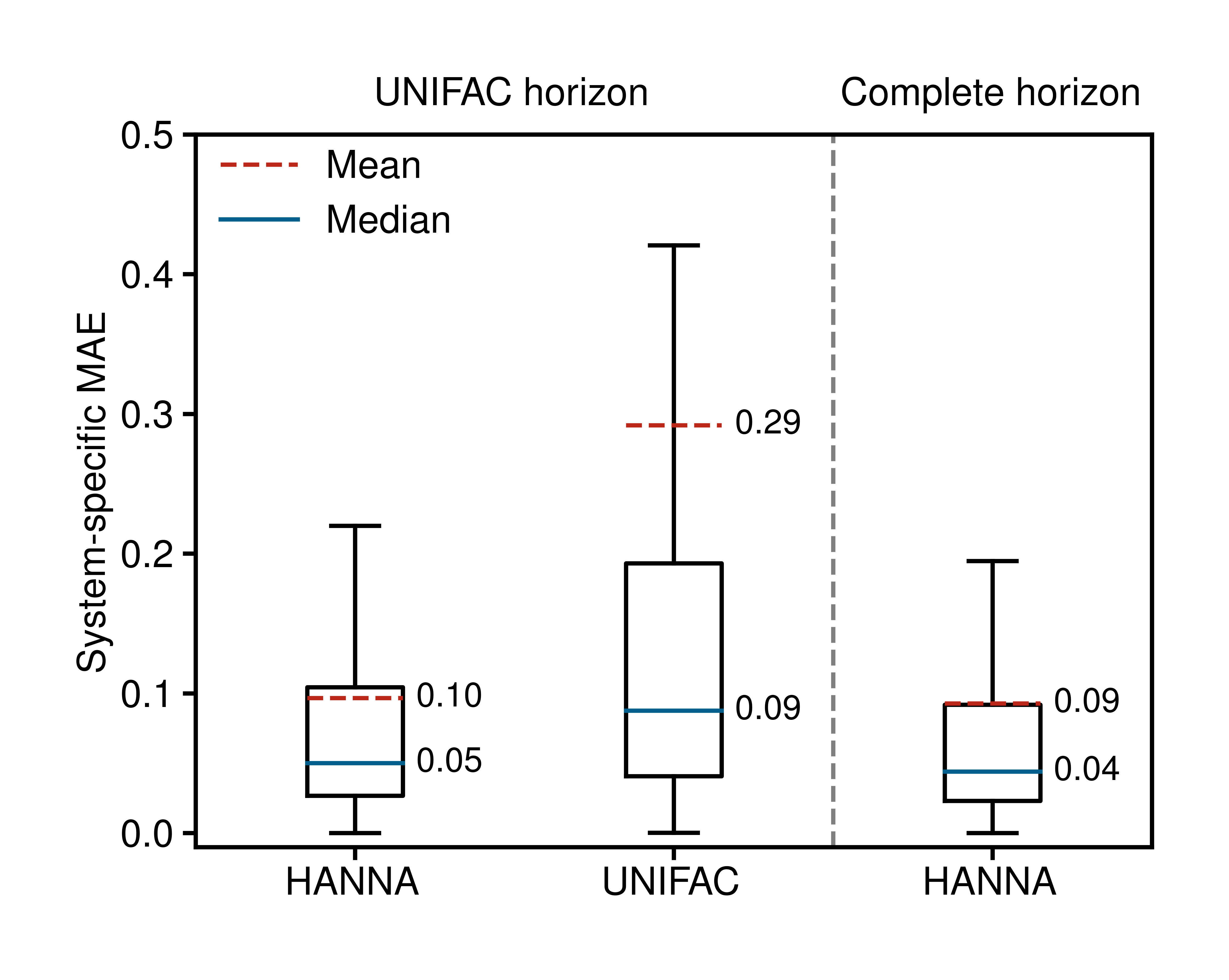}
\caption{System-specific MAE of the predicted logarithmic activity coefficients $\text{ln} \gamma_i$ from HANNA and UNIFAC. Left: results for those data from the test set that can also be predicted with UNIFAC (UNIFAC horizon). Right: results for the complete test set (complete horizon).}
\label{Figure1}
\end{figure}

Figure~\ref{Figure1} shows the system-specific MAE of the predicted logarithmic activity coefficients in boxplots; the whisker length is 1.5 times the interquartile range. Outliers are not depicted for improved visibility. The left panel of Figure~\ref{Figure1} shows the results for the UNIFAC horizon, i.e., for the data points that can be predicted with both models. HANNA significantly outperforms UNIFAC, with a mean MAE reduced to almost a third of UNIFAC's, particularly indicating a reduced number of very poorly predicted data points. Furthermore, the significantly reduced median MAE (from $0.09$ to $0.05$) indicates higher overall accuracy than UNIFAC. Figure~\ref{Figure1} (right) shows that the performance of our model on all test data (``complete horizon''), including those that cannot be predicted with UNIFAC, is similar to the UNIFAC-horizon performance. In Figure~S.4 in the Supporting Information, we show the robustness of HANNA over different random seeds for data splitting.

\FloatBarrier
As each data point in the test set corresponds to a binary system, three different cases can occur: 
\begin{enumerate}
    \item[1)] Only the combination of the two components is new, i.e., the respective system was not present in the training or validation data. However, for both components, some data (in other systems) were used for training or validation.
    \item[2)] One component is unknown, i.e., only for one of the components, some data (in other systems) were used during training or validation.
    \item[3)] Both components are unknown, i.e., no data for any of the components (in any system) were used during training or validation.
\end{enumerate}
While we do not differentiate between these cases in Figure~\ref{Figure1}, we demonstrate in Figure~S.3 in the Supporting Information that HANNA significantly outperforms UNIFAC in extrapolating to unknown components.

\begin{figure}[ht]
\centering
 \includegraphics[scale=0.47]{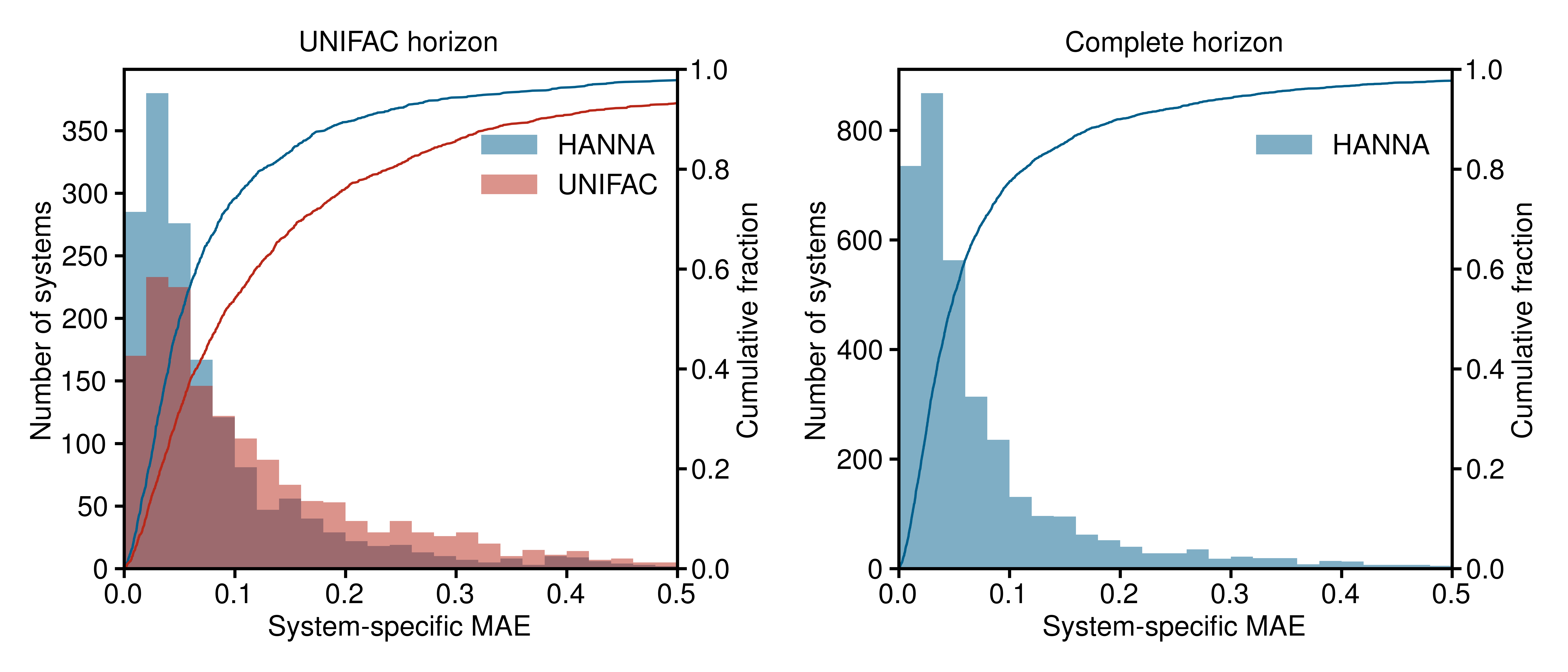}
\caption{Histograms and cumulative fractions (lines) showing the system-specific MAE for predicting logarithmic activity coefficients $\text{ln} \gamma_i$. Left: comparison of HANNA with UNIFAC on those test data that can be predicted with UNIFAC (UNIFAC horizon). The shown range covers 97.8\% of the predictions of HANNA and 93.2\% of the predictions of UNIFAC. Right: results of HANNA on the complete test set. The shown range covers 97.7\% of the predictions.}
\label{Figure2}
\end{figure}

In Figure~\ref{Figure2}, the results for the test set are shown in a histogram representation of the system-specific MAE. Furthermore, the cumulative fraction, i.e., the share of all test systems that can be predicted with an MAE smaller than the indicated value, is shown in Figure~\ref{Figure2}. Again, in the left panel, the predictions of HANNA are compared to those of UNIFAC on the UNIFAC horizon; in the right panel, the predictions of HANNA for the complete test set are shown. The results underpin the improved prediction accuracy of HANNA compared to UNIFAC, e.g., while approx. 74\% of the test systems can be predicted with $\text{MAE}<0.1$ with HANNA, which is in the range of typical experimental uncertainties for activity coefficients, this is the case for only approx. 54\% with UNIFAC.

\FloatBarrier

\begin{figure}[ht]
\centering
 \includegraphics[scale=0.31]{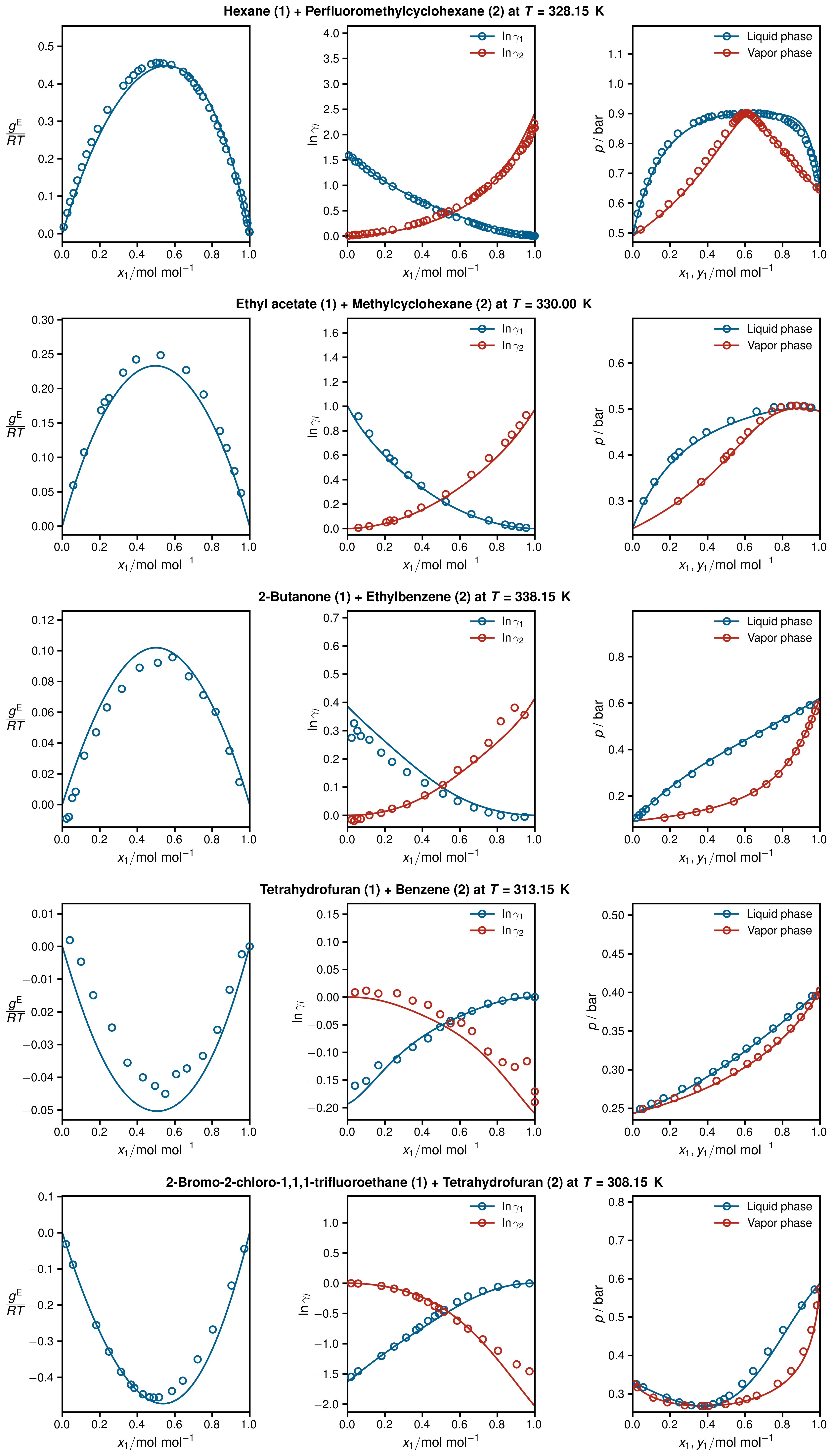}
\caption{From left to right: Gibbs excess energies $\frac{g^\text{E}}{RT}$, resulting logarithmic activity coefficients $\text{ln} \gamma_i$, and isothermal vapor-liquid phase diagrams for five systems from the test set plotted as a function of $x_1$ as predicted with HANNA (lines) and comparison to experimental test data from the DDB~\cite{DDB2023} (symbols). No data for any of the depicted systems were used for training or hyperparameter optimization.}
\label{Figure3}
\end{figure}

Figure~\ref{Figure3} shows detailed results for five isothermal systems of the test set. In addition to the predicted activity coefficients as a function of the composition of the mixtures (middle panel), the corresponding Gibbs excess energies are plotted (left panel), which are internally predicted in HANNA, cf. Figure~\ref{Overview}. Furthermore, the respective vapor-liquid phase diagrams obtained with the predicted activity coefficients are shown (right panel), cf. Section~\Newnameref{Data} for computational details. In all cases, HANNA's predictions (lines) are compared to experimental test data (symbols) from the DDB.

The shown systems were chosen randomly from the test set but to cover various phase behaviors from low-boiling azeotropes (top) over approximately ideal systems (middle) to high-boiling azeotropes (bottom). In all cases, excellent agreement is found between the predictions and the experimental data. The results also demonstrate the thermodynamic consistency of the results: $g^\text{E}=0$ and $\text{ln} \gamma_i=0$ for the pure components, and the Gibbs-Duhem equation is fulfilled throughout. 

\FloatBarrier
In Section~\textit{Ablation Studies} in the Supporting Information, results of ablation studies for which different parts in HANNA have been removed are presented. These results demonstrate the importance of hard-coding physical knowledge in the architecture of HANNA, not only regarding the thermodynamic consistency of the predictions but also regarding the predictive accuracy. Overall, the results clearly underpin the power of the hybrid approach, which combines the strengths of flexible NNs with that of physical knowledge. Given that our space of possible binary mixture is easily in the millions, even if we only take components with experimental data on activity coefficients into account, it is remarkable that HANNA can generalize well based on only a fraction of about 0.1\% of the binary systems.

\section{Conclusion}

This work introduces a novel type of thermodynamic models: a hard-constraint neural network (NN) combining the flexibility of NNs with rigorous thermodynamics. We demonstrate this for an essential thermodynamic modeling task: predicting activity coefficients in binary mixtures. The new hard-constraint hybrid NN, HANNA, incorporates thermodynamic knowledge directly into its architecture to ensure strict thermodynamic consistency. HANNA was trained end-to-end on comprehensive data from the Dortmund Data Bank (DDB).

HANNA enables thermodynamically consistent predictions for activity coefficients in any binary mixture whose components can be represented as SMILES strings. It is fully disclosed and can be used freely. The predictive capacity of HANNA was demonstrated using test data from the DDB that were not used in model development and training. HANNA clearly outperforms the best physical model for predicting activity coefficients, modified UNIFAC (Dortmund), not only in terms of prediction accuracy but also regarding the range in which it can be applied, which is basically unlimited for HANNA but restricted for UNIFAC by the availability of parameters. Only about 50 \% of the mixtures in the test data set could be modeled with UNIFAC, while all could be predicted with HANNA. 

Now that the path for developing hard-constraint NN in thermodynamics is clear, many exciting options exist. As the framework presented here is based on the Gibbs excess energy, the Gibbs-Helmholtz equation is implicitly considered so that HANNA can be extended easily to include also excess enthalpies, which is expected to improve the description of the temperature dependence of the activity coefficients. Furthermore, not only could enthalpies of mixing be incorporated, but other types of thermodynamic data could also be used, e.g., activity coefficients determined from liquid-liquid equilibria. The approach described here could also be extended to multicomponent mixtures. However, this can already be achieved by using HANNA to predict the binary subsystems and employing established physical models based on pair interaction for extrapolations to multicomponent mixtures.

Finally, the approach described here for Gibbs excess energy models can also be transferred to other thermodynamic modeling approaches, e.g., equations of state based on the Helmholtz energy. More broadly, it could be adapted to merge physical theory with NNs in other scientific fields.
\section{Methods}
\subsection{Data}
\label{Data}
Experimental data on vapor-liquid equilibria (VLE) and activity coefficients at infinite dilution in binary mixtures were taken from the Dortmund Data Bank (DDB)~\cite{DDB2023}. In preprocessing, data points labeled as poor quality by the DDB were excluded. Furthermore, only components for which a canonical SMILES string could be generated with RDKit~\cite{RDKit} from mol-files from DDB were considered.

From the VLE data, activity coefficients were calculated with extended Raoult's law:
\begin{equation}
    \gamma_{i} = \frac{p\:y_{i}}{p_i^{\text{S}}\:x_{i}}
    \label{Raoult}
\end{equation}
where $\gamma_{i}$ is the activity coefficient of component $i$ in the mixture, $x_{i}$ and $y_{i}$ are the mole fractions of component $i$ in the liquid and vapor phase in equilibrium, respectively, $p$ denotes the total pressure, and $p_i^{\text{S}}$ is the pure-component vapor pressure of $i$, which was computed using the Antoine equation with parameters from the DDB. The vapor phase was treated as a mixture of ideal gases in all cases. Furthermore, the pressure dependence of the chemical potential in the liquid phase was always neglected. Consequently, VLE data points at total pressures above 10 bar were excluded. The activity coefficients at infinite dilution, also normalized according to Raoult's law, were adopted from the DDB. The VLE diagrams in Figure~\ref{Figure3} were predicted using Equation~\eqref{Raoult} with the activity coefficients from HANNA and pure-component vapor pressures from the DDB. 

The final data set after preprocessing comprises 317,421 data points and covers 35,012 binary systems and 2,677 individual components.

\subsection{ChemBERTa-2 embeddings}
The numerical embeddings of the components were generated from a pretrained language model called ChemBERTa-2~\cite{ahmad2022}, which was trained on a large database of SMILES. We used the ``77M-MTR" model that is openly available on Huggingface~\cite{Huggingface}. The ``77M-MTR" model used 77 million SMILES to train ChemBERTa-2 in a multiregression task using the CLS token embedding~\cite{ahmad2022}. We use the CLS token embedding of the last layer of ChemBERTa-2, which results in a 384-dimensional input vector $\bm{E}_i$ for each pure component $i$, cf. Figure~\ref{Overview}. The maximum number of tokens, i.e., the individual SMILES building blocks used by ChemBERTa-2, was set to 128.  

\subsection{Data splitting, training, and evaluation of the model}
\label{Data_Splitting_Training}
For training and evaluating the hybrid model HANNA, the data set was split system-wise as follows: all data points for 80 \% of the binary systems (28,009) were used for training, all data points for another 10 \% of the systems (3,501) were used for validation and hyperparameter optimization, and all data points for the remaining 10 \% of the systems (3,502) were used to test the model. The data was randomly split, but the seed was fixed to enable reproducibility. In Figure~S.4 in the Supporting Information, we demonstrate the robustness of our approach for different seeds.

All models and training and evaluation scripts were implemented in Python 3.8.18 using PyTorch 2.1.2~\cite{NEURIPS2019_9015}. HANNA was trained on one A40 GPU using the ADAM optimizer with an initial learning rate of 0.0005, a learning rate scheduler with a decay factor of 0.1, and a patience of 10 epochs based on the validation loss. The training was stopped if the validation loss (cf. below) was not improving for 30 epochs, and the model with the best validation loss was chosen. Each training took about one hour. 

The pure-component embedding network $\bm{f}_\theta$ and the property network $f_\phi$ consist of one hidden layer, whereas the mixture embedding network $\bm{f}_\alpha$ consists of two hidden layers, cf.~Figure~\ref{Overview}. In all cases, the Sigmoid Linear Unit (SiLU) function with default PyTorch settings was used as the activation function.

HANNA uses the same number of nodes in each layer, except for the mixture embedding network $\bm{f}_\alpha$, where the input size is increased by two to include the standardized temperature and mole fraction of the respective component. Also, the output dimension of the property network $f_\phi$ is always one.

The embeddings of ChemBERTa-2 and the temperature in the training set were standardized using the StandardScaler from scikit-learn~\cite{scikit-learn}, whereas the mole fractions remained unchanged. The loss function SmoothL1Loss from PyTorch~\cite{NEURIPS2019_9015} was used to mitigate the effect of experimental outliers of the activity coefficients. The hyperparameter $\beta$ that controls the change between the L2 and L1 loss in the SmoothL1Loss was set to 0.25 and not varied. A batch size of 512 was used. The ADAM optimizer was used to update the NN weights during training. The validation data were used for early stopping, which was implemented by tracking the loss of the validation set with a patience of 30. The validation loss was also used for hyperparameter tuning. The only varied hyperparameters were the weight decay parameter $\lambda$ in the ADAM optimizer and the number of nodes in each network. Based on the results of the validation set, $\lambda=0.000001$ and 96 nodes were chosen. In the Supporting Information in Section~\textit{Hyperparameter Optimization}, we discuss the influence of the different hyperparameters and present the validation loss results. 

We provide a ``final'' version of HANNA with this paper that was trained as described above, except that no test set was used, i.e., 90 \% of all systems were used for training and 10 \% for validation.

\section*{Acknowledgments}
The authors gratefully acknowledge financial support by Carl Zeiss Foundation in the frame of the project ‘Process Engineering 4.0’ and by DFG in the frame of the Priority Program SPP2363 'Molecular Machine Learning' (grant number 497201843). Furthermore, FJ gratefully acknowledges financial support by DFG in the frame of the Emmy-Noether program (grant number 528649696). SM acknowledges support from the National Science Foundation (NSF) under an NSF CAREER Award, award numbers 2003237 and 2007719, by the Department of Energy under grant DE-SC0022331, and by the IARPA WRIVA program.
\section*{Data availability}
All data were taken from the Dortmund Data Bank~\cite{DDB2023}. The final version of HANNA, which was trained and validated on 100\% of the data  (without using a test set), is available on Github (https://github.com/tspecht93/HANNA) and distributed under the MIT license.
\section*{Supplementary information}
Supplementary material with additional results and technical details is available.
\section*{Competing interests}
The authors declare no competing interests.

\begin{mcitethebibliography}{68}
\providecommand*\natexlab[1]{#1}
\providecommand*\mciteSetBstSublistMode[1]{}
\providecommand*\mciteSetBstMaxWidthForm[2]{}
\providecommand*\mciteBstWouldAddEndPuncttrue
  {\def\EndOfBibitem{\unskip.}}
\providecommand*\mciteBstWouldAddEndPunctfalse
  {\let\EndOfBibitem\relax}
\providecommand*\mciteSetBstMidEndSepPunct[3]{}
\providecommand*\mciteSetBstSublistLabelBeginEnd[3]{}
\providecommand*\EndOfBibitem{}
\mciteSetBstSublistMode{f}
\mciteSetBstMaxWidthForm{subitem}{(\alph{mcitesubitemcount})}
\mciteSetBstSublistLabelBeginEnd
  {\mcitemaxwidthsubitemform\space}
  {\relax}
  {\relax}

\bibitem[Ravindran(2022)]{Ravindran2022}
Ravindran,~S. Five ways deep learning has transformed image analysis. \emph{Nature} \textbf{2022}, \emph{609}, 864–866, DOI: \doi{10.1038/d41586-022-02964-6}\relax
\mciteBstWouldAddEndPuncttrue
\mciteSetBstMidEndSepPunct{\mcitedefaultmidpunct}
{\mcitedefaultendpunct}{\mcitedefaultseppunct}\relax
\EndOfBibitem
\bibitem[LeCun \latin{et~al.}(2015)LeCun, Bengio, and Hinton]{LeCun2015}
LeCun,~Y.; Bengio,~Y.; Hinton,~G. Deep learning. \emph{Nature} \textbf{2015}, \emph{521}, 436–444, DOI: \doi{10.1038/nature14539}\relax
\mciteBstWouldAddEndPuncttrue
\mciteSetBstMidEndSepPunct{\mcitedefaultmidpunct}
{\mcitedefaultendpunct}{\mcitedefaultseppunct}\relax
\EndOfBibitem
\bibitem[Jumper \latin{et~al.}(2021)Jumper, Evans, Pritzel, Green, Figurnov, Ronneberger, Tunyasuvunakool, Bates, Žídek, Potapenko, Bridgland, Meyer, Kohl, Ballard, Cowie, Romera-Paredes, Nikolov, Jain, Adler, Back, Petersen, Reiman, Clancy, Zielinski, Steinegger, Pacholska, Berghammer, Bodenstein, Silver, Vinyals, Senior, Kavukcuoglu, Kohli, and Hassabis]{Jumper2021}
Jumper,~J. \latin{et~al.}  Highly accurate protein structure prediction with AlphaFold. \emph{Nature} \textbf{2021}, \emph{596}, 583–589, DOI: \doi{10.1038/s41586-021-03819-2}\relax
\mciteBstWouldAddEndPuncttrue
\mciteSetBstMidEndSepPunct{\mcitedefaultmidpunct}
{\mcitedefaultendpunct}{\mcitedefaultseppunct}\relax
\EndOfBibitem
\bibitem[Abramson \latin{et~al.}(2024)Abramson, Adler, Dunger, Evans, Green, Pritzel, Ronneberger, Willmore, Ballard, Bambrick, Bodenstein, Evans, Hung, O’Neill, Reiman, Tunyasuvunakool, Wu, Žemgulytė, Arvaniti, Beattie, Bertolli, Bridgland, Cherepanov, Congreve, Cowen-Rivers, Cowie, Figurnov, Fuchs, Gladman, Jain, Khan, Low, Perlin, Potapenko, Savy, Singh, Stecula, Thillaisundaram, Tong, Yakneen, Zhong, Zielinski, Žídek, Bapst, Kohli, Jaderberg, Hassabis, and Jumper]{Abramson2024}
Abramson,~J. \latin{et~al.}  Accurate structure prediction of biomolecular interactions with AlphaFold 3. \emph{Nature} \textbf{2024}, \emph{630}, 493–500, DOI: \doi{10.1038/s41586-024-07487-w}\relax
\mciteBstWouldAddEndPuncttrue
\mciteSetBstMidEndSepPunct{\mcitedefaultmidpunct}
{\mcitedefaultendpunct}{\mcitedefaultseppunct}\relax
\EndOfBibitem
\bibitem[M.~Bran \latin{et~al.}(2024)M.~Bran, Cox, Schilter, Baldassari, White, and Schwaller]{MBran2024}
M.~Bran,~A. \latin{et~al.}  Augmenting large language models with chemistry tools. \emph{Nature Machine Intelligence} \textbf{2024}, \emph{6}, 525–535, DOI: \doi{10.1038/s42256-024-00832-8}\relax
\mciteBstWouldAddEndPuncttrue
\mciteSetBstMidEndSepPunct{\mcitedefaultmidpunct}
{\mcitedefaultendpunct}{\mcitedefaultseppunct}\relax
\EndOfBibitem
\bibitem[Van~Veen \latin{et~al.}(2024)Van~Veen, Van~Uden, Blankemeier, Delbrouck, Aali, Bluethgen, Pareek, Polacin, Reis, Seehofnerová, Rohatgi, Hosamani, Collins, Ahuja, Langlotz, Hom, Gatidis, Pauly, and Chaudhari]{VanVeen2024}
Van~Veen,~D. \latin{et~al.}  Adapted large language models can outperform medical experts in clinical text summarization. \emph{Nature Medicine} \textbf{2024}, \emph{30}, 1134–1142, DOI: \doi{10.1038/s41591-024-02855-5}\relax
\mciteBstWouldAddEndPuncttrue
\mciteSetBstMidEndSepPunct{\mcitedefaultmidpunct}
{\mcitedefaultendpunct}{\mcitedefaultseppunct}\relax
\EndOfBibitem
\bibitem[Hornik \latin{et~al.}(1989)Hornik, Stinchcombe, and White]{Hornik1989}
Hornik,~K.; Stinchcombe,~M.; White,~H. Multilayer feedforward networks are universal approximators. \emph{Neural Networks} \textbf{1989}, \emph{2}, 359–366, DOI: \doi{10.1016/0893-6080(89)90020-8}\relax
\mciteBstWouldAddEndPuncttrue
\mciteSetBstMidEndSepPunct{\mcitedefaultmidpunct}
{\mcitedefaultendpunct}{\mcitedefaultseppunct}\relax
\EndOfBibitem
\bibitem[Venkatasubramanian(2018)]{Venkatasubramanian2018}
Venkatasubramanian,~V. The promise of artificial intelligence in chemical engineering: Is it here, finally? \emph{AIChE Journal} \textbf{2018}, \emph{65}, 466–478, DOI: \doi{10.1002/aic.16489}\relax
\mciteBstWouldAddEndPuncttrue
\mciteSetBstMidEndSepPunct{\mcitedefaultmidpunct}
{\mcitedefaultendpunct}{\mcitedefaultseppunct}\relax
\EndOfBibitem
\bibitem[Fang \latin{et~al.}(2022)Fang, Liu, Lei, He, Zhang, Zhou, Wang, Wu, and Wang]{Fang2022}
Fang,~X. \latin{et~al.}  Geometry-enhanced molecular representation learning for property prediction. \emph{Nature Machine Intelligence} \textbf{2022}, \emph{4}, 127–134, DOI: \doi{10.1038/s42256-021-00438-4}\relax
\mciteBstWouldAddEndPuncttrue
\mciteSetBstMidEndSepPunct{\mcitedefaultmidpunct}
{\mcitedefaultendpunct}{\mcitedefaultseppunct}\relax
\EndOfBibitem
\bibitem[Li \latin{et~al.}(2022)Li, Wang, Lee, and Luo]{Li2022}
Li,~R. \latin{et~al.}  Physics-informed deep learning for solving phonon Boltzmann transport equation with large temperature non-equilibrium. \emph{npj Computational Materials} \textbf{2022}, \emph{8}, DOI: \doi{10.1038/s41524-022-00712-y}\relax
\mciteBstWouldAddEndPuncttrue
\mciteSetBstMidEndSepPunct{\mcitedefaultmidpunct}
{\mcitedefaultendpunct}{\mcitedefaultseppunct}\relax
\EndOfBibitem
\bibitem[Zang \latin{et~al.}(2023)Zang, Zhao, and Tang]{Zang2023}
Zang,~X.; Zhao,~X.; Tang,~B. Hierarchical molecular graph self-supervised learning for property prediction. \emph{Communications Chemistry} \textbf{2023}, \emph{6}, DOI: \doi{10.1038/s42004-023-00825-5}\relax
\mciteBstWouldAddEndPuncttrue
\mciteSetBstMidEndSepPunct{\mcitedefaultmidpunct}
{\mcitedefaultendpunct}{\mcitedefaultseppunct}\relax
\EndOfBibitem
\bibitem[Schweidtmann(2024)]{Schweidtmann2024}
Schweidtmann,~A.~M. Generative artificial intelligence in chemical engineering. \emph{Nature Chemical Engineering} \textbf{2024}, \emph{1}, 193–193, DOI: \doi{10.1038/s44286-024-00041-5}\relax
\mciteBstWouldAddEndPuncttrue
\mciteSetBstMidEndSepPunct{\mcitedefaultmidpunct}
{\mcitedefaultendpunct}{\mcitedefaultseppunct}\relax
\EndOfBibitem
\bibitem[Rittig \latin{et~al.}(2023)Rittig, Felton, Lapkin, and Mitsos]{Rittig2023}
Rittig,~J.~G. \latin{et~al.}  Gibbs–Duhem-informed neural networks for binary activity coefficient prediction. \emph{Digital Discovery} \textbf{2023}, \emph{2}, 1752–1767, DOI: \doi{10.1039/d3dd00103b}\relax
\mciteBstWouldAddEndPuncttrue
\mciteSetBstMidEndSepPunct{\mcitedefaultmidpunct}
{\mcitedefaultendpunct}{\mcitedefaultseppunct}\relax
\EndOfBibitem
\bibitem[Raissi \latin{et~al.}(2019)Raissi, Perdikaris, and Karniadakis]{Raissi2019}
Raissi,~M.; Perdikaris,~P.; Karniadakis,~G. Physics-informed neural networks: A deep learning framework for solving forward and inverse problems involving nonlinear partial differential equations. \emph{Journal of Computational Physics} \textbf{2019}, \emph{378}, 686–707, DOI: \doi{10.1016/j.jcp.2018.10.045}\relax
\mciteBstWouldAddEndPuncttrue
\mciteSetBstMidEndSepPunct{\mcitedefaultmidpunct}
{\mcitedefaultendpunct}{\mcitedefaultseppunct}\relax
\EndOfBibitem
\bibitem[Lin \latin{et~al.}(2022)Lin, Wang, and Zhang]{Lin2022}
Lin,~G.; Wang,~Y.; Zhang,~Z. Multi-variance replica exchange SGMCMC for inverse and forward problems via Bayesian PINN. \emph{Journal of Computational Physics} \textbf{2022}, \emph{460}, 111173, DOI: \doi{10.1016/j.jcp.2022.111173}\relax
\mciteBstWouldAddEndPuncttrue
\mciteSetBstMidEndSepPunct{\mcitedefaultmidpunct}
{\mcitedefaultendpunct}{\mcitedefaultseppunct}\relax
\EndOfBibitem
\bibitem[Zhu \latin{et~al.}(2022)Zhu, Khademi, Charalampidis, and Kevrekidis]{Zhu2021}
Zhu,~W. \latin{et~al.}  Neural networks enforcing physical symmetries in nonlinear dynamical lattices: The case example of the Ablowitz–Ladik model. \emph{Physica D: Nonlinear Phenomena} \textbf{2022}, \emph{434}, 133264, DOI: \doi{10.1016/j.physd.2022.133264}\relax
\mciteBstWouldAddEndPuncttrue
\mciteSetBstMidEndSepPunct{\mcitedefaultmidpunct}
{\mcitedefaultendpunct}{\mcitedefaultseppunct}\relax
\EndOfBibitem
\bibitem[Molnar and Grauer(2022)Molnar, and Grauer]{Molnar2022}
Molnar,~J.~P.; Grauer,~S.~J. Flow field tomography with uncertainty quantification using a Bayesian physics-informed neural network. \emph{Measurement Science and Technology} \textbf{2022}, \emph{33}, 065305, DOI: \doi{10.1088/1361-6501/ac5437}\relax
\mciteBstWouldAddEndPuncttrue
\mciteSetBstMidEndSepPunct{\mcitedefaultmidpunct}
{\mcitedefaultendpunct}{\mcitedefaultseppunct}\relax
\EndOfBibitem
\bibitem[Psaros \latin{et~al.}(2022)Psaros, Kawaguchi, and Karniadakis]{Psaros2021}
Psaros,~A.~F.; Kawaguchi,~K.; Karniadakis,~G.~E. Meta-learning PINN loss functions. \emph{Journal of Computational Physics} \textbf{2022}, \emph{458}, 111121, DOI: \doi{10.1016/j.jcp.2022.111121}\relax
\mciteBstWouldAddEndPuncttrue
\mciteSetBstMidEndSepPunct{\mcitedefaultmidpunct}
{\mcitedefaultendpunct}{\mcitedefaultseppunct}\relax
\EndOfBibitem
\bibitem[Martin and Schaub(2022)Martin, and Schaub]{Martin2022}
Martin,~J.; Schaub,~H. Physics-informed neural networks for gravity field modeling of the Earth and Moon. \emph{Celestial Mechanics and Dynamical Astronomy} \textbf{2022}, \emph{134}, DOI: \doi{10.1007/s10569-022-10069-5}\relax
\mciteBstWouldAddEndPuncttrue
\mciteSetBstMidEndSepPunct{\mcitedefaultmidpunct}
{\mcitedefaultendpunct}{\mcitedefaultseppunct}\relax
\EndOfBibitem
\bibitem[Zhao \latin{et~al.}(2022)Zhao, Stuebner, Lua, Phan, and Yan]{Zhao2022}
Zhao,~Z. \latin{et~al.}  Full-field temperature recovery during water quenching processes via physics-informed machine learning. \emph{Journal of Materials Processing Technology} \textbf{2022}, \emph{303}, 117534, DOI: \doi{10.1016/j.jmatprotec.2022.117534}\relax
\mciteBstWouldAddEndPuncttrue
\mciteSetBstMidEndSepPunct{\mcitedefaultmidpunct}
{\mcitedefaultendpunct}{\mcitedefaultseppunct}\relax
\EndOfBibitem
\bibitem[Karniadakis \latin{et~al.}(2021)Karniadakis, Kevrekidis, Lu, Perdikaris, Wang, and Yang]{Karniadakis2021}
Karniadakis,~G.~E. \latin{et~al.}  Physics-informed machine learning. \emph{Nature Reviews Physics} \textbf{2021}, \emph{3}, 422–440, DOI: \doi{10.1038/s42254-021-00314-5}\relax
\mciteBstWouldAddEndPuncttrue
\mciteSetBstMidEndSepPunct{\mcitedefaultmidpunct}
{\mcitedefaultendpunct}{\mcitedefaultseppunct}\relax
\EndOfBibitem
\bibitem[Xu and Darve(2020)Xu, and Darve]{Xu2020}
Xu,~K.; Darve,~E. Physics constrained learning for data-driven inverse modeling from sparse observations. \emph{arXiv:2002.10521} \textbf{2020}, DOI: \doi{10.48550/ARXIV.2002.10521}\relax
\mciteBstWouldAddEndPuncttrue
\mciteSetBstMidEndSepPunct{\mcitedefaultmidpunct}
{\mcitedefaultendpunct}{\mcitedefaultseppunct}\relax
\EndOfBibitem
\bibitem[Chen \latin{et~al.}(2021)Chen, Huang, Zhang, Zeng, Wang, Zhang, and Yan]{Chen2021}
Chen,~Y. \latin{et~al.}  Theory-guided hard constraint projection (HCP): A knowledge-based data-driven scientific machine learning method. \emph{Journal of Computational Physics} \textbf{2021}, \emph{445}, 110624, DOI: \doi{10.1016/j.jcp.2021.110624}\relax
\mciteBstWouldAddEndPuncttrue
\mciteSetBstMidEndSepPunct{\mcitedefaultmidpunct}
{\mcitedefaultendpunct}{\mcitedefaultseppunct}\relax
\EndOfBibitem
\bibitem[Chen and Zhang(2020)Chen, and Zhang]{Chen2020}
Chen,~Y.; Zhang,~D. Physics-constrained indirect supervised learning. \emph{Theoretical and Applied Mechanics Letters} \textbf{2020}, \emph{10}, 155–160, DOI: \doi{10.1016/j.taml.2020.01.019}\relax
\mciteBstWouldAddEndPuncttrue
\mciteSetBstMidEndSepPunct{\mcitedefaultmidpunct}
{\mcitedefaultendpunct}{\mcitedefaultseppunct}\relax
\EndOfBibitem
\bibitem[Márquez-Neila \latin{et~al.}(2017)Márquez-Neila, Salzmann, and Fua]{Neila2017}
Márquez-Neila,~P.; Salzmann,~M.; Fua,~P. Imposing hard constraints on deep networks: promises and limitations. \emph{arXiv:1706.02025} \textbf{2017}, DOI: \doi{10.48550/ARXIV.1706.02025}\relax
\mciteBstWouldAddEndPuncttrue
\mciteSetBstMidEndSepPunct{\mcitedefaultmidpunct}
{\mcitedefaultendpunct}{\mcitedefaultseppunct}\relax
\EndOfBibitem
\bibitem[Lu \latin{et~al.}(2021)Lu, Pestourie, Yao, Wang, Verdugo, and Johnson]{Lu2021}
Lu,~L. \latin{et~al.}  Physics-informed neural networks with hard constraints for inverse design. \emph{SIAM Journal on Scientific Computing} \textbf{2021}, \emph{43}, B1105–B1132, DOI: \doi{10.1137/21m1397908}\relax
\mciteBstWouldAddEndPuncttrue
\mciteSetBstMidEndSepPunct{\mcitedefaultmidpunct}
{\mcitedefaultendpunct}{\mcitedefaultseppunct}\relax
\EndOfBibitem
\bibitem[Gmehling \latin{et~al.}(2019)Gmehling, Kleiber, Kolbe, and Rarey]{Gmehling2019-vx}
Gmehling,~J. \latin{et~al.}  \emph{Chemical Thermodynamics for Process Simulation}, 2nd ed.; John Wiley \& Sons, 2019\relax
\mciteBstWouldAddEndPuncttrue
\mciteSetBstMidEndSepPunct{\mcitedefaultmidpunct}
{\mcitedefaultendpunct}{\mcitedefaultseppunct}\relax
\EndOfBibitem
\bibitem[Renon and Prausnitz(1968)Renon, and Prausnitz]{Renon1968}
Renon,~H.; Prausnitz,~J.~M. Local compositions in thermodynamic excess functions for liquid mixtures. \emph{AIChE Journal} \textbf{1968}, \emph{14}, 135–144, DOI: \doi{10.1002/aic.690140124}\relax
\mciteBstWouldAddEndPuncttrue
\mciteSetBstMidEndSepPunct{\mcitedefaultmidpunct}
{\mcitedefaultendpunct}{\mcitedefaultseppunct}\relax
\EndOfBibitem
\bibitem[Abrams and Prausnitz(1975)Abrams, and Prausnitz]{Abrams1975}
Abrams,~D.~S.; Prausnitz,~J.~M. Statistical thermodynamics of liquid mixtures: A new expression for the excess Gibbs energy of partly or completely miscible systems. \emph{AIChE Journal} \textbf{1975}, \emph{21}, 116–128, DOI: \doi{10.1002/aic.690210115}\relax
\mciteBstWouldAddEndPuncttrue
\mciteSetBstMidEndSepPunct{\mcitedefaultmidpunct}
{\mcitedefaultendpunct}{\mcitedefaultseppunct}\relax
\EndOfBibitem
\bibitem[Klamt(1995)]{Klamt1995}
Klamt,~A. Conductor-like screening model for real solvents: A new approach to the quantitative calculation of solvation phenomena. \emph{The Journal of Physical Chemistry} \textbf{1995}, \emph{99}, 2224–2235, DOI: \doi{10.1021/j100007a062}\relax
\mciteBstWouldAddEndPuncttrue
\mciteSetBstMidEndSepPunct{\mcitedefaultmidpunct}
{\mcitedefaultendpunct}{\mcitedefaultseppunct}\relax
\EndOfBibitem
\bibitem[Lin and Sandler(2001)Lin, and Sandler]{Lin2001}
Lin,~S.-T.; Sandler,~S.~I. A priori phase equilibrium prediction from a segment contribution solvation model. \emph{Industrial \& Engineering Chemistry Research} \textbf{2001}, \emph{41}, 899–913, DOI: \doi{10.1021/ie001047w}\relax
\mciteBstWouldAddEndPuncttrue
\mciteSetBstMidEndSepPunct{\mcitedefaultmidpunct}
{\mcitedefaultendpunct}{\mcitedefaultseppunct}\relax
\EndOfBibitem
\bibitem[Hsieh \latin{et~al.}(2010)Hsieh, Sandler, and Lin]{Hsieh2010}
Hsieh,~C.-M.; Sandler,~S.~I.; Lin,~S.-T. Improvements of COSMO-SAC for vapor–liquid and liquid–liquid equilibrium predictions. \emph{Fluid Phase Equilibria} \textbf{2010}, \emph{297}, 90–97, DOI: \doi{10.1016/j.fluid.2010.06.011}\relax
\mciteBstWouldAddEndPuncttrue
\mciteSetBstMidEndSepPunct{\mcitedefaultmidpunct}
{\mcitedefaultendpunct}{\mcitedefaultseppunct}\relax
\EndOfBibitem
\bibitem[Hsieh \latin{et~al.}(2014)Hsieh, Lin, and Vrabec]{Hsieh2014}
Hsieh,~C.-M.; Lin,~S.-T.; Vrabec,~J. Considering the dispersive interactions in the COSMO-SAC model for more accurate predictions of fluid phase behavior. \emph{Fluid Phase Equilibria} \textbf{2014}, \emph{367}, 109–116, DOI: \doi{10.1016/j.fluid.2014.01.032}\relax
\mciteBstWouldAddEndPuncttrue
\mciteSetBstMidEndSepPunct{\mcitedefaultmidpunct}
{\mcitedefaultendpunct}{\mcitedefaultseppunct}\relax
\EndOfBibitem
\bibitem[Fredenslund \latin{et~al.}(1975)Fredenslund, Jones, and Prausnitz]{Fredenslund1975}
Fredenslund,~A.; Jones,~R.~L.; Prausnitz,~J.~M. Group‐contribution estimation of activity coefficients in nonideal liquid mixtures. \emph{AIChE Journal} \textbf{1975}, \emph{21}, 1086–1099, DOI: \doi{10.1002/aic.690210607}\relax
\mciteBstWouldAddEndPuncttrue
\mciteSetBstMidEndSepPunct{\mcitedefaultmidpunct}
{\mcitedefaultendpunct}{\mcitedefaultseppunct}\relax
\EndOfBibitem
\bibitem[Weidlich and Gmehling(1987)Weidlich, and Gmehling]{Weidlich1987}
Weidlich,~U.; Gmehling,~J. A modified UNIFAC model. 1. Prediction of VLE, $h^\text{E}$, and $\gamma^{\infty}$. \emph{Industrial \& Engineering Chemistry Research} \textbf{1987}, \emph{26}, 1372–1381, DOI: \doi{10.1021/ie00067a018}\relax
\mciteBstWouldAddEndPuncttrue
\mciteSetBstMidEndSepPunct{\mcitedefaultmidpunct}
{\mcitedefaultendpunct}{\mcitedefaultseppunct}\relax
\EndOfBibitem
\bibitem[Fingerhut \latin{et~al.}(2017)Fingerhut, Chen, Schedemann, Cordes, Rarey, Hsieh, Vrabec, and Lin]{Fingerhut2017}
Fingerhut,~R. \latin{et~al.}  Comprehensive assessment of COSMO-SAC models for predictions of fluid-phase equilibria. \emph{Industrial \& Engineering Chemistry Research} \textbf{2017}, \emph{56}, 9868–9884, DOI: \doi{10.1021/acs.iecr.7b01360}\relax
\mciteBstWouldAddEndPuncttrue
\mciteSetBstMidEndSepPunct{\mcitedefaultmidpunct}
{\mcitedefaultendpunct}{\mcitedefaultseppunct}\relax
\EndOfBibitem
\bibitem[Jirasek \latin{et~al.}(2020)Jirasek, Bamler, and Mandt]{Jirasek2020b}
Jirasek,~F.; Bamler,~R.; Mandt,~S. Hybridizing physical and data-driven prediction methods for physicochemical properties. \emph{Chemical Communications} \textbf{2020}, \emph{56}, 12407–12410, DOI: \doi{10.1039/d0cc05258b}\relax
\mciteBstWouldAddEndPuncttrue
\mciteSetBstMidEndSepPunct{\mcitedefaultmidpunct}
{\mcitedefaultendpunct}{\mcitedefaultseppunct}\relax
\EndOfBibitem
\bibitem[Jirasek \latin{et~al.}(2020)Jirasek, Alves, Damay, Vandermeulen, Bamler, Bortz, Mandt, Kloft, and Hasse]{Jirasek2020a}
Jirasek,~F. \latin{et~al.}  Machine learning in thermodynamics: prediction of activity coefficients by matrix completion. \emph{The Journal of Physical Chemistry Letters} \textbf{2020}, \emph{11}, 981–985, DOI: \doi{10.1021/acs.jpclett.9b03657}\relax
\mciteBstWouldAddEndPuncttrue
\mciteSetBstMidEndSepPunct{\mcitedefaultmidpunct}
{\mcitedefaultendpunct}{\mcitedefaultseppunct}\relax
\EndOfBibitem
\bibitem[Damay \latin{et~al.}(2023)Damay, Ryzhakov, Jirasek, Hasse, Oseledets, and Bortz]{Damay2023}
Damay,~J. \latin{et~al.}  Predicting temperature‐dependent activity coefficients at infinite dilution using tensor completion. \emph{Chemie Ingenieur Technik} \textbf{2023}, \emph{95}, 1061–1069, DOI: \doi{10.1002/cite.202200230}\relax
\mciteBstWouldAddEndPuncttrue
\mciteSetBstMidEndSepPunct{\mcitedefaultmidpunct}
{\mcitedefaultendpunct}{\mcitedefaultseppunct}\relax
\EndOfBibitem
\bibitem[Sanchez~Medina \latin{et~al.}(2022)Sanchez~Medina, Linke, Stoll, and Sundmacher]{SanchezMedina2022}
Sanchez~Medina,~E.~I. \latin{et~al.}  Graph neural networks for the prediction of infinite dilution activity coefficients. \emph{Digital Discovery} \textbf{2022}, \emph{1}, 216–225, DOI: \doi{10.1039/d1dd00037c}\relax
\mciteBstWouldAddEndPuncttrue
\mciteSetBstMidEndSepPunct{\mcitedefaultmidpunct}
{\mcitedefaultendpunct}{\mcitedefaultseppunct}\relax
\EndOfBibitem
\bibitem[Santana \latin{et~al.}(2024)Santana, Rebello, Queiroz, Ribeiro, Shardt, and Nogueira]{Santana2024}
Santana,~V.~V. \latin{et~al.}  PUFFIN: A path-unifying feed-forward interfaced network for vapor pressure prediction. \emph{Chemical Engineering Science} \textbf{2024}, \emph{286}, 119623, DOI: \doi{10.1016/j.ces.2023.119623}\relax
\mciteBstWouldAddEndPuncttrue
\mciteSetBstMidEndSepPunct{\mcitedefaultmidpunct}
{\mcitedefaultendpunct}{\mcitedefaultseppunct}\relax
\EndOfBibitem
\bibitem[Habicht \latin{et~al.}(2023)Habicht, Brandenbusch, and Sadowski]{Habicht2023}
Habicht,~J.; Brandenbusch,~C.; Sadowski,~G. Predicting PC-SAFT pure-component parameters by machine learning using a molecular fingerprint as key input. \emph{Fluid Phase Equilibria} \textbf{2023}, \emph{565}, 113657, DOI: \doi{10.1016/j.fluid.2022.113657}\relax
\mciteBstWouldAddEndPuncttrue
\mciteSetBstMidEndSepPunct{\mcitedefaultmidpunct}
{\mcitedefaultendpunct}{\mcitedefaultseppunct}\relax
\EndOfBibitem
\bibitem[Deng \latin{et~al.}(2023)Deng, Yang, Wang, Ojima, Samaras, and Wang]{Deng2023}
Deng,~J. \latin{et~al.}  A systematic study of key elements underlying molecular property prediction. \emph{Nature Communications} \textbf{2023}, \emph{14}, DOI: \doi{10.1038/s41467-023-41948-6}\relax
\mciteBstWouldAddEndPuncttrue
\mciteSetBstMidEndSepPunct{\mcitedefaultmidpunct}
{\mcitedefaultendpunct}{\mcitedefaultseppunct}\relax
\EndOfBibitem
\bibitem[Shilpa \latin{et~al.}(2023)Shilpa, Kashyap, and Sunoj]{Shilpa2023}
Shilpa,~S.; Kashyap,~G.; Sunoj,~R.~B. Recent applications of machine learning in molecular property and chemical reaction outcome predictions. \emph{The Journal of Physical Chemistry A} \textbf{2023}, \emph{127}, 8253–8271, DOI: \doi{10.1021/acs.jpca.3c04779}\relax
\mciteBstWouldAddEndPuncttrue
\mciteSetBstMidEndSepPunct{\mcitedefaultmidpunct}
{\mcitedefaultendpunct}{\mcitedefaultseppunct}\relax
\EndOfBibitem
\bibitem[Aouichaoui \latin{et~al.}(2023)Aouichaoui, Fan, Mansouri, Abildskov, and Sin]{Aouichaoui2023}
Aouichaoui,~A. R.~N. \latin{et~al.}  Combining group-contribution concept and graph neural networks toward interpretable molecular property models. \emph{Journal of Chemical Information and Modeling} \textbf{2023}, \emph{63}, 725–744, DOI: \doi{10.1021/acs.jcim.2c01091}\relax
\mciteBstWouldAddEndPuncttrue
\mciteSetBstMidEndSepPunct{\mcitedefaultmidpunct}
{\mcitedefaultendpunct}{\mcitedefaultseppunct}\relax
\EndOfBibitem
\bibitem[Hayer \latin{et~al.}(2022)Hayer, Jirasek, and Hasse]{Hayer2022}
Hayer,~N.; Jirasek,~F.; Hasse,~H. Prediction of Henry’s law constants by matrix completion. \emph{AIChE Journal} \textbf{2022}, \emph{68}, DOI: \doi{10.1002/aic.17753}\relax
\mciteBstWouldAddEndPuncttrue
\mciteSetBstMidEndSepPunct{\mcitedefaultmidpunct}
{\mcitedefaultendpunct}{\mcitedefaultseppunct}\relax
\EndOfBibitem
\bibitem[Großmann \latin{et~al.}(2022)Großmann, Bellaire, Hayer, Jirasek, and Hasse]{Gromann2022}
Großmann,~O. \latin{et~al.}  Database for liquid phase diffusion coefficients at infinite dilution at 298 K and matrix completion methods for their prediction. \emph{Digital Discovery} \textbf{2022}, \emph{1}, 886–897, DOI: \doi{10.1039/d2dd00073c}\relax
\mciteBstWouldAddEndPuncttrue
\mciteSetBstMidEndSepPunct{\mcitedefaultmidpunct}
{\mcitedefaultendpunct}{\mcitedefaultseppunct}\relax
\EndOfBibitem
\bibitem[Jirasek and Hasse(2023)Jirasek, and Hasse]{Jirasek2023Review}
Jirasek,~F.; Hasse,~H. Combining machine learning with physical knowledge in thermodynamic modeling of fluid mixtures. \emph{Annual Review of Chemical and Biomolecular Engineering} \textbf{2023}, \emph{14}, 31–51, DOI: \doi{10.1146/annurev-chembioeng-092220-025342}\relax
\mciteBstWouldAddEndPuncttrue
\mciteSetBstMidEndSepPunct{\mcitedefaultmidpunct}
{\mcitedefaultendpunct}{\mcitedefaultseppunct}\relax
\EndOfBibitem
\bibitem[Winter \latin{et~al.}(2022)Winter, Winter, Schilling, and Bardow]{Winter2022}
Winter,~B. \latin{et~al.}  A smile is all you need: predicting limiting activity coefficients from SMILES with natural language processing. \emph{Digital Discovery} \textbf{2022}, \emph{1}, 859–869, DOI: \doi{10.1039/d2dd00058j}\relax
\mciteBstWouldAddEndPuncttrue
\mciteSetBstMidEndSepPunct{\mcitedefaultmidpunct}
{\mcitedefaultendpunct}{\mcitedefaultseppunct}\relax
\EndOfBibitem
\bibitem[Jirasek \latin{et~al.}(2022)Jirasek, Bamler, Fellenz, Bortz, Kloft, Mandt, and Hasse]{Jirasek2022}
Jirasek,~F. \latin{et~al.}  Making thermodynamic models of mixtures predictive by machine learning: matrix completion of pair interactions. \emph{Chemical Science} \textbf{2022}, \emph{13}, 4854–4862, DOI: \doi{10.1039/d1sc07210b}\relax
\mciteBstWouldAddEndPuncttrue
\mciteSetBstMidEndSepPunct{\mcitedefaultmidpunct}
{\mcitedefaultendpunct}{\mcitedefaultseppunct}\relax
\EndOfBibitem
\bibitem[Jirasek \latin{et~al.}(2023)Jirasek, Hayer, Abbas, Schmid, and Hasse]{Jirasek2023_MCM}
Jirasek,~F. \latin{et~al.}  Prediction of parameters of group contribution models of mixtures by matrix completion. \emph{Physical Chemistry Chemical Physics} \textbf{2023}, \emph{25}, 1054–1062, DOI: \doi{10.1039/d2cp04478a}\relax
\mciteBstWouldAddEndPuncttrue
\mciteSetBstMidEndSepPunct{\mcitedefaultmidpunct}
{\mcitedefaultendpunct}{\mcitedefaultseppunct}\relax
\EndOfBibitem
\bibitem[Winter \latin{et~al.}(2023)Winter, Winter, Esper, Schilling, and Bardow]{Winter2023}
Winter,~B. \latin{et~al.}  SPT-NRTL: A physics-guided machine learning model to predict thermodynamically consistent activity coefficients. \emph{Fluid Phase Equilibria} \textbf{2023}, \emph{568}, 113731, DOI: \doi{10.1016/j.fluid.2023.113731}\relax
\mciteBstWouldAddEndPuncttrue
\mciteSetBstMidEndSepPunct{\mcitedefaultmidpunct}
{\mcitedefaultendpunct}{\mcitedefaultseppunct}\relax
\EndOfBibitem
\bibitem[Werner \latin{et~al.}(2023)Werner, Seidel, Jafar, Heese, Hasse, and Bortz]{Werner2023}
Werner,~J. \latin{et~al.}  Multiplicities in thermodynamic activity coefficients. \emph{AIChE Journal} \textbf{2023}, \emph{69}, DOI: \doi{10.1002/aic.18251}\relax
\mciteBstWouldAddEndPuncttrue
\mciteSetBstMidEndSepPunct{\mcitedefaultmidpunct}
{\mcitedefaultendpunct}{\mcitedefaultseppunct}\relax
\EndOfBibitem
\bibitem[Rarey(2005)]{Rarey2005}
Rarey,~J. Extended flexibility for $G^\text{E}$ models and simultaneous description of vapor-liquid equilibrium and liquid-liquid equilibrium using a nonlinear transformation of the concentration dependence. \emph{Industrial \& Engineering Chemistry Research} \textbf{2005}, \emph{44}, 7600–7608, DOI: \doi{10.1021/ie050431w}\relax
\mciteBstWouldAddEndPuncttrue
\mciteSetBstMidEndSepPunct{\mcitedefaultmidpunct}
{\mcitedefaultendpunct}{\mcitedefaultseppunct}\relax
\EndOfBibitem
\bibitem[Marcilla \latin{et~al.}(2011)Marcilla, Olaya, and Serrano]{Marcilla2011}
Marcilla,~A.; Olaya,~M.~M.; Serrano,~M.~D. Liquid-vapor equilibrium data correlation: part I. Pitfalls and some ideas to overcome them. \emph{Industrial \& Engineering Chemistry Research} \textbf{2011}, \emph{50}, 4077–4085, DOI: \doi{10.1021/ie101909d}\relax
\mciteBstWouldAddEndPuncttrue
\mciteSetBstMidEndSepPunct{\mcitedefaultmidpunct}
{\mcitedefaultendpunct}{\mcitedefaultseppunct}\relax
\EndOfBibitem
\bibitem[Marcilla \latin{et~al.}(2018)Marcilla, Olaya, and Reyes-Labarta]{Marcilla2018}
Marcilla,~A.; Olaya,~M.; Reyes-Labarta,~J. The unavoidable necessity of considering temperature dependence of the liquid Gibbs energy of mixing for certain VLE data correlations. \emph{Fluid Phase Equilibria} \textbf{2018}, \emph{473}, 17–31, DOI: \doi{10.1016/j.fluid.2018.05.025}\relax
\mciteBstWouldAddEndPuncttrue
\mciteSetBstMidEndSepPunct{\mcitedefaultmidpunct}
{\mcitedefaultendpunct}{\mcitedefaultseppunct}\relax
\EndOfBibitem
\bibitem[Marcilla \latin{et~al.}(2019)Marcilla, Olaya, Reyes-Labarta, and Carbonell-Hermida]{Marcilla2019}
Marcilla,~A. \latin{et~al.}  Procedure for the correlation of normal appearance VLE data, where the classical models dramatically fail with no apparent reason. \emph{Fluid Phase Equilibria} \textbf{2019}, \emph{493}, 88–101, DOI: \doi{10.1016/j.fluid.2019.04.001}\relax
\mciteBstWouldAddEndPuncttrue
\mciteSetBstMidEndSepPunct{\mcitedefaultmidpunct}
{\mcitedefaultendpunct}{\mcitedefaultseppunct}\relax
\EndOfBibitem
\bibitem[Weininger(1988)]{Weininger1988}
Weininger,~D. SMILES, a chemical language and information system. 1. Introduction to methodology and encoding rules. \emph{Journal of Chemical Information and Computer Sciences} \textbf{1988}, \emph{28}, 31–36, DOI: \doi{10.1021/ci00057a005}\relax
\mciteBstWouldAddEndPuncttrue
\mciteSetBstMidEndSepPunct{\mcitedefaultmidpunct}
{\mcitedefaultendpunct}{\mcitedefaultseppunct}\relax
\EndOfBibitem
\bibitem[Ahmad \latin{et~al.}(2022)Ahmad, Simon, Chithrananda, Grand, and Ramsundar]{ahmad2022}
Ahmad,~W. \latin{et~al.}  ChemBERTa-2: towards chemical foundation models. \emph{arXiv:2209.01712} \textbf{2022}, DOI: \doi{10.48550/ARXIV.2209.01712}\relax
\mciteBstWouldAddEndPuncttrue
\mciteSetBstMidEndSepPunct{\mcitedefaultmidpunct}
{\mcitedefaultendpunct}{\mcitedefaultseppunct}\relax
\EndOfBibitem
\bibitem[Zaheer \latin{et~al.}(2017)Zaheer, Kottur, Ravanbakhsh, Poczos, Salakhutdinov, and Smola]{Zaheer2017}
Zaheer,~M. \latin{et~al.}  Deep sets. \emph{arXiv:1703.06114} \textbf{2017}, DOI: \doi{10.48550/ARXIV.1703.06114}\relax
\mciteBstWouldAddEndPuncttrue
\mciteSetBstMidEndSepPunct{\mcitedefaultmidpunct}
{\mcitedefaultendpunct}{\mcitedefaultseppunct}\relax
\EndOfBibitem
\bibitem[Hanaoka(2020)]{Hanaoka2020}
Hanaoka,~K. Deep Neural Networks for Multicomponent Molecular Systems. \emph{ACS Omega} \textbf{2020}, \emph{5}, 21042–21053, DOI: \doi{10.1021/acsomega.0c02599}\relax
\mciteBstWouldAddEndPuncttrue
\mciteSetBstMidEndSepPunct{\mcitedefaultmidpunct}
{\mcitedefaultendpunct}{\mcitedefaultseppunct}\relax
\EndOfBibitem
\bibitem[Deiters and Kraska(2012)Deiters, and Kraska]{Deiters2012-me}
Deiters,~U.~K.; Kraska,~T. \emph{High-Pressure Fluid Phase Equilibria}, 1st ed.; Elsevier, 2012\relax
\mciteBstWouldAddEndPuncttrue
\mciteSetBstMidEndSepPunct{\mcitedefaultmidpunct}
{\mcitedefaultendpunct}{\mcitedefaultseppunct}\relax
\EndOfBibitem
\bibitem[Paszke \latin{et~al.}(2019)Paszke, Gross, Massa, Lerer, Bradbury, Chanan, Killeen, Lin, Gimelshein, Antiga, Desmaison, K\"{o}pf, Yang, DeVito, Raison, Tejani, Chilamkurthy, Steiner, Fang, Bai, and Chintala]{NEURIPS2019_9015}
Paszke,~A. \latin{et~al.}  PyTorch: An imperative style, high-performance deep learning library. \emph{arXiv:1912.01703} \textbf{2019}, DOI: \doi{10.48550/ARXIV.1912.01703}\relax
\mciteBstWouldAddEndPuncttrue
\mciteSetBstMidEndSepPunct{\mcitedefaultmidpunct}
{\mcitedefaultendpunct}{\mcitedefaultseppunct}\relax
\EndOfBibitem
\bibitem[DDB(2023)]{DDB2023}
Dortmund Data Bank. www.ddbst.com, 2023\relax
\mciteBstWouldAddEndPuncttrue
\mciteSetBstMidEndSepPunct{\mcitedefaultmidpunct}
{\mcitedefaultendpunct}{\mcitedefaultseppunct}\relax
\EndOfBibitem
\bibitem[RDK()]{RDKit}
RDKit: Open-source cheminformatics. (Last accessed: 04.04.2024), \url{https://www.rdkit.org}\relax
\mciteBstWouldAddEndPuncttrue
\mciteSetBstMidEndSepPunct{\mcitedefaultmidpunct}
{\mcitedefaultendpunct}{\mcitedefaultseppunct}\relax
\EndOfBibitem
\bibitem[Hug()]{Huggingface}
Huggingface ChemBERTa-2 model. https://huggingface.co/DeepChem/ChemBERTa-77M-MTR, Last accessed: 05.04.2024\relax
\mciteBstWouldAddEndPuncttrue
\mciteSetBstMidEndSepPunct{\mcitedefaultmidpunct}
{\mcitedefaultendpunct}{\mcitedefaultseppunct}\relax
\EndOfBibitem
\bibitem[Pedregosa \latin{et~al.}(2011)Pedregosa, Varoquaux, Gramfort, Michel, Thirion, Grisel, Blondel, Prettenhofer, Weiss, Dubourg, Vanderplas, Passos, Cournapeau, Brucher, Perrot, and Duchesnay]{scikit-learn}
Pedregosa,~F. \latin{et~al.}  Scikit-learn: machine learning in python. \emph{Journal of Machine Learning Research} \textbf{2011}, \emph{12}, 2825--2830\relax
\mciteBstWouldAddEndPuncttrue
\mciteSetBstMidEndSepPunct{\mcitedefaultmidpunct}
{\mcitedefaultendpunct}{\mcitedefaultseppunct}\relax
\EndOfBibitem
\end{mcitethebibliography}
\providecommand{\latin}[1]{#1}
\makeatletter
\providecommand{\doi}
  {\begingroup\let\do\@makeother\dospecials
  \catcode`\{=1 \catcode`\}=2 \doi@aux}
\providecommand{\doi@aux}[1]{\endgroup\texttt{#1}}
\makeatother
\providecommand*\mcitethebibliography{\thebibliography}
\csname @ifundefined\endcsname{endmcitethebibliography}  {\let\endmcitethebibliography\endthebibliography}{}

\end{document}


\renewcommand{\thesection}{S.\arabic{section}}
\renewcommand{\theequation}{S.\arabic{equation}}
\renewcommand{\thefigure}{S.\arabic{figure}}
\renewcommand{\thetable}{S.\arabic{table}}



\section{Ablation studies}
\label{Ablation_Study}
\begin{figure}[ht]
\centering
 \includegraphics[scale=0.6]{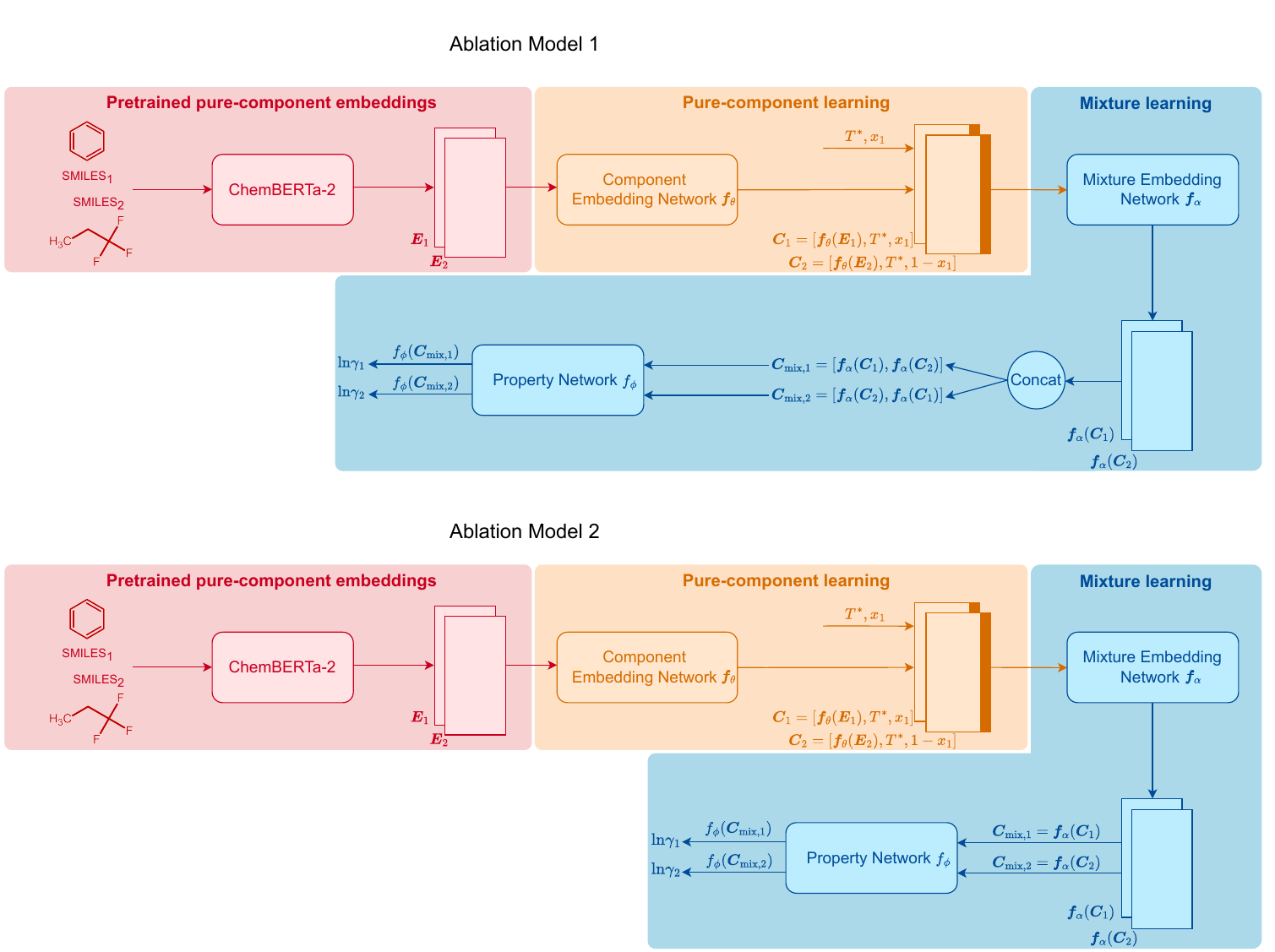}
\caption{Architecture of the ablation models.}
\label{FigureSI1}
\end{figure}

This section discusses the results of two ablation experiments to understand the importance of different parts of HANNA's architecture. Figure~\ref{FigureSI1} shows the architecture of the ablation models. In ablation model 1, the logarithmic activity coefficients $\text{ln} \gamma_i$ are predicted directly without considering the thermodynamic constraints. We had to adapt the architecture slightly further to ensure an interaction modeling between the components. Therefore, the mixture embeddings $\bm{f}_\alpha(\bm{C}_1)$ and $\bm{f}_\alpha(\bm{C}_2)$ are concatenated to $\bm{C}_{\text{mix},1}=[\bm{f}_\alpha(\bm{C}_1),\bm{f}_\alpha(\bm{C}_2)]$ and $\bm{C}_{\text{mix},2}=[\bm{f}_\alpha(\bm{C}_2),\bm{f}_\alpha(\bm{C}_1)]$. Each of them is then processed through the property network $f_\phi$ to calculate $\text{ln} \gamma_1$ and $\text{ln} \gamma_2$, respectively.

In ablation model 2, the logarithmic activity coefficients  $\text{ln} \gamma_i$ are again modeled directly without the intermediate prediction of $g^\text{E}$. However, we now use the mixture embedding $\bm{f}_\alpha(\bm{C}_i)$ to predict $\text{ln} \gamma_i$ in the property network $f_\phi$. Note that the activity coefficient is still modeled as an equivariant property in both ablation models due to our deep-set architecture.

\FloatBarrier

\begin{figure}[ht]
\centering
 \includegraphics[scale=0.37]{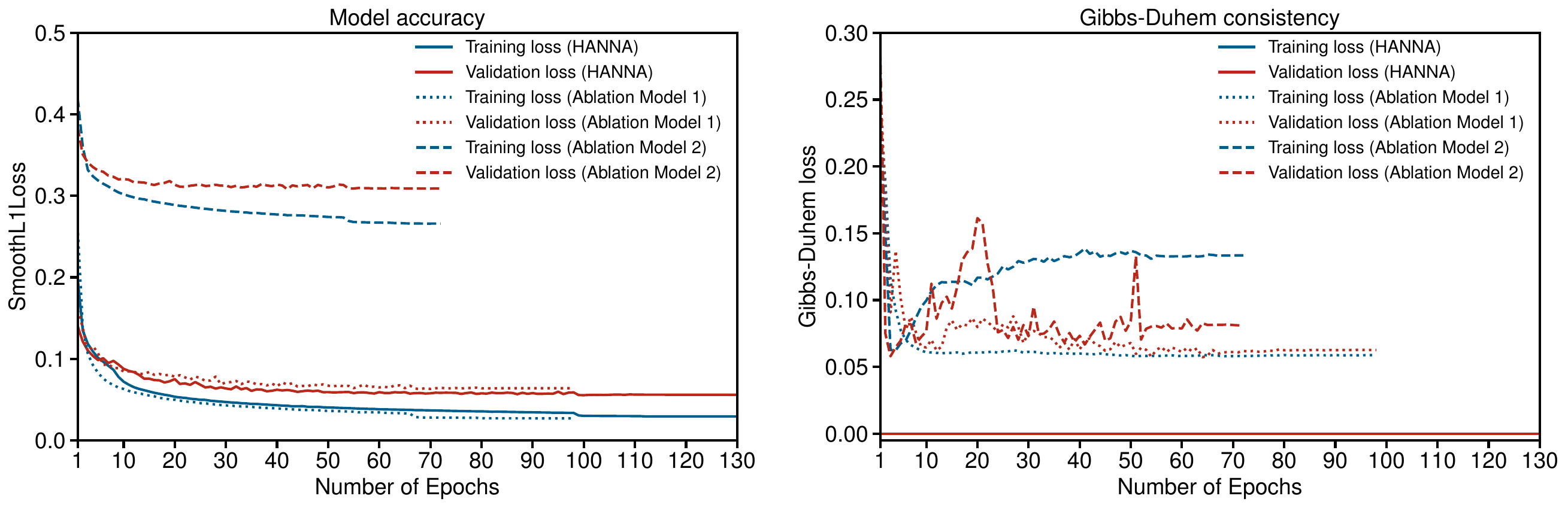}
\caption{Left: SmoothL1Loss for training and validation set over the epochs for HANNA, ablation model 1, and ablation model 2. Right: Mean squared deviation of model predictions from the Gibbs-Duhem equation, cf.~Equation~(2) in the manuscript for training and validation set over the epochs, for the three models. The end of the training is determined by the validation loss, cf. Section~\textit{Data splitting, training and evaluation of model} in the manuscript for details.}
\label{FigureSI2}
\end{figure}

Figure~\ref{FigureSI2} (left) shows the training and validation SmoothL1Loss over the epochs for HANNA, ablation model 1, and ablation model 2. Ablation model 1 shows nearly the same loss as HANNA, whereas ablation model 2 shows a significant score deterioration. These results underpin that interaction modeling is necessary, either through a concatenation of the features of the components as in ablation model 1 or through a summation as in HANNA. Furthermore, the results show that the model flexibility is not overly restricted by hard-constraining its predictions to the thermodynamically consistent solution space.

Figure~\ref{FigureSI2} (right) shows the mean squared deviation of the model predictions from the Gibbs-Duhem equation, cf.~Equation~(2) in the manuscript,  for training and validation set over the epochs for HANNA, ablation model 1, and ablation model 2. As expected, HANNA shows a consistent error of zero over all epochs since Gibbs-Duhem consistency is strictly enforced in the network architecture. In contrast, strict Gibbs-Duhem consistency is not obtained with the ablations models. Specifically, the loss of ablation model 1 decreases during the first epochs; the model obviously ``learns'' something about the Gibbs-Duhem equation during the training, but only until a certain threshold is reached. We can assume that this is, among others, caused by inconsistencies in the experimental training data~\cite{Gmehling2019-vx}. Hence, learning Gibbs-Duhem consistency solely based on experimental data seems unfeasible. Ablation model 2 also learns something about Gibbs-Duhem consistency during the first epochs, but then it shows instabilities. 

\FloatBarrier

\section{Extrapolation to unknown components}
\label{Unknown components}
\begin{figure}[ht]
\centering
 \includegraphics[scale=0.66]{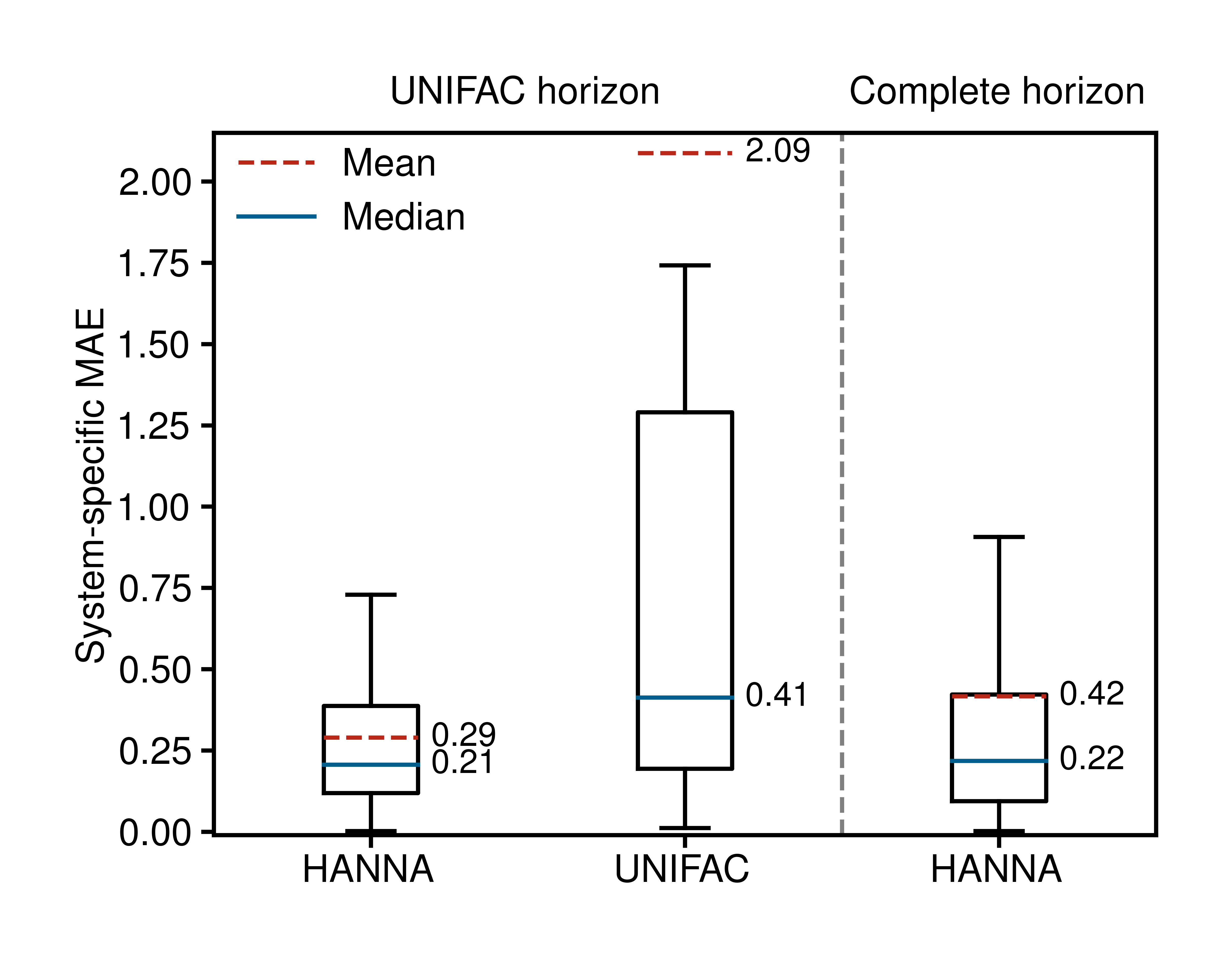}
\caption{
System-specific MAE of the predicted logarithmic activity coefficients $\text{ln} \gamma_i$ from HANNA and UNIFAC for the case that one component is unknown in the system. Left: results for the 40 out of 1658 systems from the test set that can also be predicted with UNIFAC (UNIFAC horizon). Right: results for the 56 out of 3502 systems from the complete test set (complete horizon).}
\label{FigureSI3}
\end{figure}

Figure~\ref{FigureSI3} compares HANNA and UNIFAC for predicting the systems from the test set where one component is entirely unknown to HANNA. Furthermore, the results from HANNA for all systems with one unknown component from the complete test set (complete horizon) are shown. As discussed in the manuscript, there could, in principle, be another class of test systems in which both components are unknown to HANNA. However, since this was the case for precisely a single system within the UNIFAC horizon, the respective results are omitted here. Note that we can distinguish these cases only for HANNA since the training set of UNIFAC has not been disclosed.

The boxplots demonstrate that HANNA is significantly more reliable in extrapolating to systems containing unknown components than UNIFAC. For both models, the scores shown here are significantly worse than those shown in Figure~2 of the manuscript, which also covers the test data points where only the system was unknown to HANNA. This observation might be explained considering two facts: first, systems containing water are heavily overrepresented in the data set shown in Figure~\ref{FigureSI3} (70\% of the systems contain water), and systems with water are known to be rather tricky to describe. Second, data for activity coefficients at infinite dilution are heavily over-represented in the data set shown in Figure~\ref{FigureSI3} (80\% of the systems), which are, again, more difficult to predict than data at finite concentrations, among others, due to the high experimental uncertainty of activity coefficients at infinite dilution.

\FloatBarrier

\section{Hyperparameter optimization}
\label{Hyperparameter_Study}
Table~\ref{tab:hyperparameters} shows the varied hyperparameters in developing HANNA and the SmoothL1Loss achieved on the validation data. Model 5 was used throughout the manuscript.
\begin{table}[ht]
\centering
\caption{Varied hyperparameters in the development of HANNA and their influence on the validation loss. $\lambda$ is the weight decay in the ADAM optimizer.}
\begin{tabular}{@{}llll@{}}

\toprule
Model No. & $\lambda$ & Number of Nodes & SmoothL1Loss \\ \midrule
1         &  0.0000001      &  128 &  0.0576\\
2         &   0.000001     &    128& 0.0576  \\
3         &   0.00001     &    128 &  0.0588\\
4         &   0.0000001     &   96 &  0.0557 \\
5         &    0.000001    &    96 & 0.0554 \\
6         &      0.00001  &     96 & 0.0571\\
7         &   0.0000001     &   64 &  0.0607\\
8         &    0.000001    &    64&   0.0606\\
9         &     0.00001   &     64 & 0.0621\\ \bottomrule
\end{tabular}

\label{tab:hyperparameters}
\end{table}
\FloatBarrier

\section{Results for different seeds}
\label{Seed_Study}
\begin{figure}[ht]
\centering
 \includegraphics[scale=0.5]{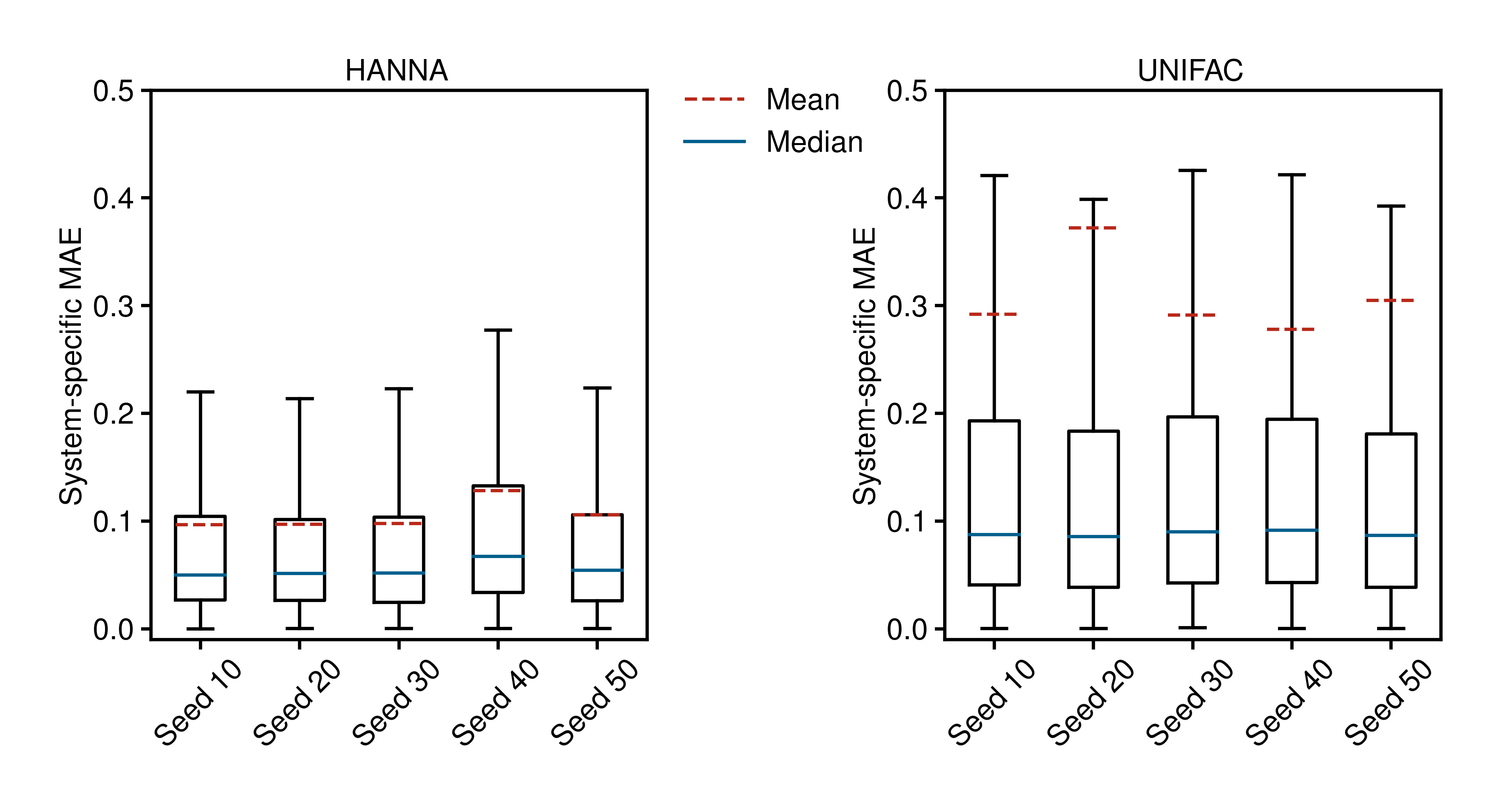}
\caption{System-specific MAE of HANNA (left) and UNIFAC (right) on the UNIFAC horizon for different test sets, defined by specifying different seeds in the data split. Seed 10 was used throughout the manuscript.}
\label{FigureSI4}
\end{figure}

This section compares results for different splittings of the systems from the complete dataset into training, validation, and test sets. Figure~\ref{FigureSI4} shows boxplots for the system-specific MAE on the different test sets, indicated by different specified seeds, on the UNIFAC horizon. High robustness of HANNA is observed.
\FloatBarrier


\providecommand{\latin}[1]{#1}
\makeatletter
\providecommand{\doi}
  {\begingroup\let\do\@makeother\dospecials
  \catcode`\{=1 \catcode`\}=2 \doi@aux}
\providecommand{\doi@aux}[1]{\endgroup\texttt{#1}}
\makeatother
\providecommand*\mcitethebibliography{\thebibliography}
\csname @ifundefined\endcsname{endmcitethebibliography}  {\let\endmcitethebibliography\endthebibliography}{}